\documentclass{article}
\ifdefined\pdfminorversion\pdfminorversion=7\fi
\PassOptionsToPackage{dvipsnames}{xcolor}

\usepackage{tikz}

\providecommand{\dartworkshoptitle}{Foundations of LLM Post-Training in Changing Environments}
\usepackage[dblblindworkshop,final]{neurips_2026}
\workshoptitle{\dartworkshoptitle}

\usepackage[utf8]{inputenc}
\usepackage[T1]{fontenc}
\usepackage{hyperref}
\hypersetup{colorlinks=true,linkcolor=black,citecolor=black,urlcolor=black}
\hypersetup{
  pdftitle={Diverse Alternative Reasoning Traces for Hard Math Problems},
  pdfauthor={Lukas Aichberger, Adam Golinski, Arno Blaas, Arwen Bradley, Jacob Springer, Mehrdad Farajtabar, Michael Kirchhof, Sinead Williamson}
}
\usepackage{url}
\usepackage{caption}
\usepackage{subcaption}
\usepackage{multirow}
\usepackage{booktabs}
\usepackage{amsmath,amsfonts}
\usepackage{algorithm}
\usepackage{algpseudocode}
\algrenewcommand\algorithmicrequire{\textbf{Input:}}
\algrenewcommand\algorithmicensure{\textbf{Output:}}
\algnewcommand{\IfThen}[2]{\State \algorithmicif\ #1\ \algorithmicthen\ #2}
\usepackage{cleveref}
\usepackage{adjustbox}
\usepackage{latexsym}
\usepackage{microtype}
\usepackage{inconsolata}
\usepackage{graphicx}
\usepackage{wrapfig}

\usepackage{enumitem}
\setitemize{noitemsep,topsep=0pt,parsep=0pt,partopsep=0pt}
\usepackage{fvextra}
\DefineVerbatimEnvironment{prompttext}{Verbatim}{%
  fontsize=\small,frame=none,breaklines,commandchars=\\\{\}}
\usepackage{xcolor}
\usepackage[skins,breakable]{tcolorbox}
\usepackage{setspace}
\usepackage{comment}
\usepackage{mathtools}

\newcommand{\method}{\texttt{DART}}
\newcommand{\fullmethod}{\texttt{Diverse Alternative Reasoning Traces}}

\usetikzlibrary{patterns}
\tcbset{
  striped box/.style={
    enhanced,
    frame hidden,
    boxrule=0pt,
    colback=white,
    underlay={
      \pattern[pattern=horizontal lines, pattern color=gray,
        pattern preaction={fill=white}]
        (frame.south west) rectangle (frame.north east);
    },
    borderline west={1pt}{0pt}{black},
    borderline east={1pt}{0pt}{black},
    boxsep=5pt,
    left=5pt,
    right=5pt,
  }
}

\newtcolorbox{promptbox}[1][]{%
  enhanced,
  breakable,
  colback=gray!5!white,
  colframe=gray!85!black,
  boxrule=0.5pt,
  arc=1mm,
  left=2mm,
  right=2mm,
  top=2mm,
  bottom=2mm,
  coltitle=black,
  colbacktitle=black!15!white,
  lefttitle=2mm,
  righttitle=2mm,
  toptitle=2mm,
  bottomtitle=2mm,
  #1
}

\definecolor{AppleBlue3}{RGB}{169,202,223}
\definecolor{AppleBlue5}{RGB}{71,130,166}
\definecolor{AppleGreen4}{RGB}{126,193,155}
\definecolor{AppleRed4}{RGB}{234,135,143}
\definecolor{AppleCoolGray}{RGB}{116,128,139}
\definecolor{AppleBlue}{RGB}{84,151,193}
\definecolor{AppleGreen}{RGB}{83,172,121}

\newtcolorbox{reasoningbox}[1][]{%
  enhanced,
  breakable,
  colback=black!5!white,
  colframe=black!85!white,
  boxrule=0.5pt,
  arc=1mm,
  left=2mm,
  right=2mm,
  top=2mm,
  bottom=2mm,
  coltitle=black,
  colbacktitle=black!15!white,
  lefttitle=2mm,
  righttitle=2mm,
  toptitle=2mm,
  bottomtitle=2mm,
  #1
}

\newlist{reasonsteps}{itemize}{1}
\setlist[reasonsteps]{
  label=\texttt{[STEP]},
  leftmargin=*,
  itemsep=8pt,
  align=parleft
}

\crefformat{equation}{Eq.~#2\textup{#1}#3}
\crefformat{figure}{Fig.~#2\textup{#1}#3}
\crefformat{table}{Tab.~#2\textup{#1}#3}
\crefformat{appendix}{App.~#2\textup{#1}#3}
\crefformat{section}{\S#2#1#3}
\crefmultiformat{section}{\S\S#2#1#3}{ and~#2#1#3}{, #2#1#3}{, and~#2#1#3}
\crefmultiformat{figure}{Figs.~#2#1#3}{ and~#2#1#3}{, #2#1#3}{, and~#2#1#3}
\crefmultiformat{table}{Tabs.~#2#1#3}{ and~#2#1#3}{, #2#1#3}{, and~#2#1#3}
\crefmultiformat{appendix}{Apps.~#2#1#3}{ and~#2#1#3}{, #2#1#3}{, and~#2#1#3}
\crefrangeformat{appendix}{Apps.~#3#1#4--#5#2#6}
\crefrangeformat{figure}{Figs.~#3#1#4--#5#2#6}
\crefrangeformat{section}{\S\S#3#1#4--#5#2#6}

\DeclareRobustCommand{\colorswatch}[1]{%
  \textcolor{#1}{\rule{0.8em}{0.5em}}%
}
\DeclareRobustCommand{\crossedswatch}[1]{%
  \tikz[baseline=0pt]{%
    \draw[#1] (0,0) rectangle (0.8em,0.5em);
    \draw[#1,thick] (0,0) -- (0.8em,0.5em);
  }%
}

\title{Diverse Alternative Reasoning Traces \\for Hard Math Problems}

\author{
\textbf{Lukas Aichberger} \thanks{\ Johannes Kepler University Linz (Work done
during an internship at Apple Inc.)} \ , \
\textbf{Adam Goli{\'n}ski} \thanks{\ Apple Inc. Correspondence to: Sinead Williamson \texttt{<sa\_williamson@apple.com>}} \ , \
\textbf{Arno Blaas} \footnotemark[2] \ , \
\textbf{Arwen Bradley} \footnotemark[2] \ , \
\textbf{Jacob Springer} \thanks{\ Carnegie Mellon University (Work done
during an internship at Apple Inc.)} \ , \\
\ \textbf{Mehrdad Farajtabar} \footnotemark[2] \ , \
\textbf{Michael Kirchhof} \footnotemark[2] \ , \
\textbf{Sinead Williamson} \footnotemark[2]
}

\makeatletter\let\@fnsymbol\@arabic\makeatother

\begin{document}
\maketitle

\vspace{-2pt}
\begin{abstract}
Post-training large language models (LLMs) for mathematical reasoning often relies on accepted traces whose final answers pass a verifier. For hard math problems, however, such traces can be rare under independent Monte Carlo (MC) sampling.
To address this, we propose \fullmethod~(\method), a trace-mining method that starts from a rejected trace, decomposes it into reasoning steps, and ranks them by position-weighted uncertainty. It then generates deliberately distinct alternative candidates for highly ranked steps using reference-answer guidance and a meta-thought before completing the resulting traces.
On greedy-unsolved DeepMath-103K problems, \method{} recovers 37\% more problems than MC sampling at a matched number of generated traces. Its estimated generation cost per recovered problem is nearly identical, while its cost per accepted trace is 30\% lower.
Post-training Llama-3.1-8B-Instruct on traces mined by \method{} achieves the highest point estimate in five of six evaluations across MATH-500 and AIME benchmarks.
These results demonstrate that revisiting rejected traces with \method{} can expand problem coverage and provide effective post-training data.
\end{abstract}

\section{Introduction}
\label{sec:introduction}

Large language models (LLMs) are increasingly employed in tasks that require complex, multi-step problem solving \citep{srivastava2023}.
One such task is mathematical reasoning, where prompting LLMs to generate intermediate reasoning steps rather than directly predicting the final answer has been shown to substantially improve their problem-solving performance \citep{weiChainofThoughtPromptingElicits2023}.

\begin{figure}[t]
    \centering
    \includegraphics[width=\linewidth,trim=0cm 17.8cm 17.2cm 0cm,clip]{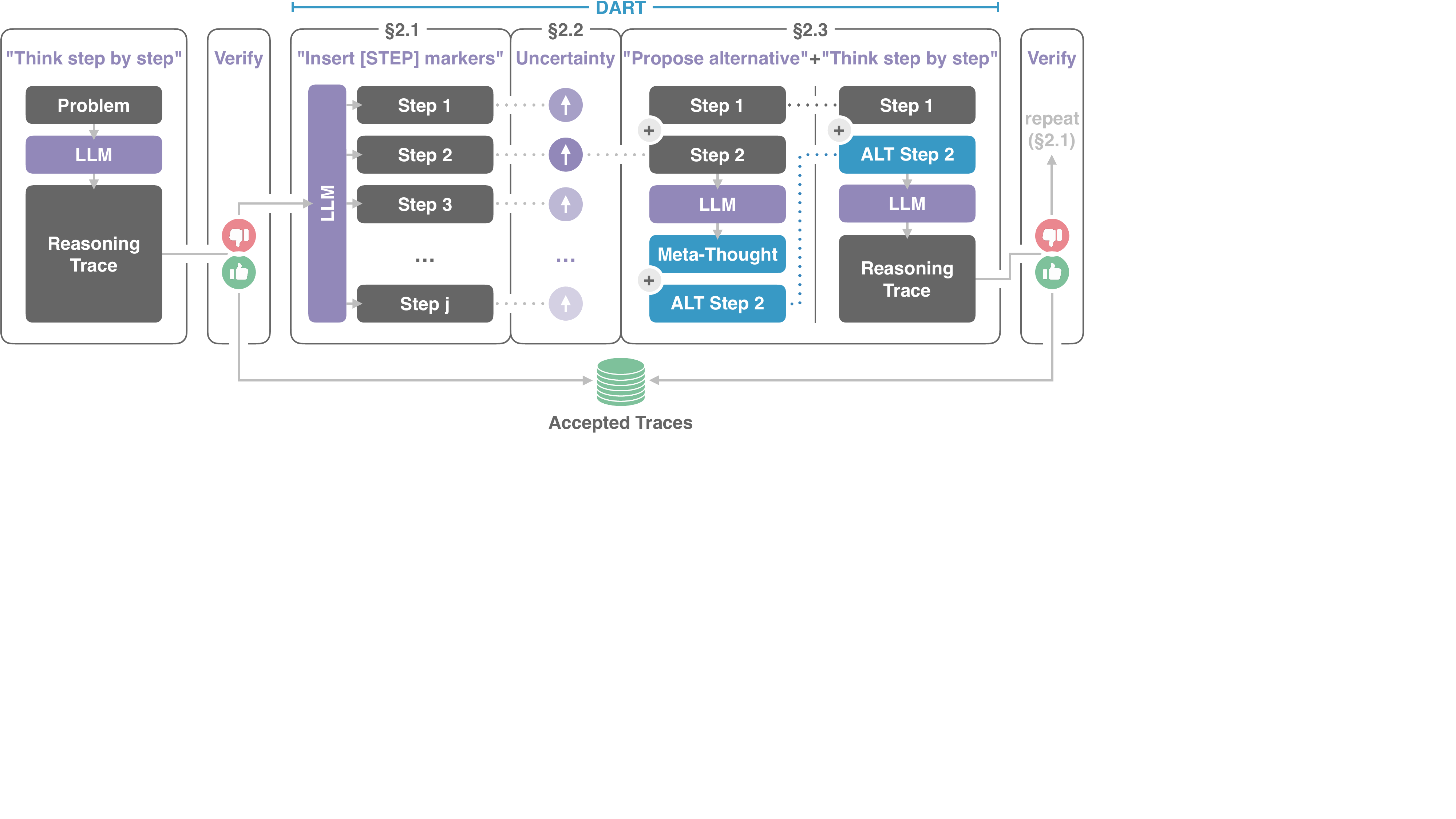}
    \caption{
    \method{} decomposes a rejected trace into self-contained reasoning steps (\Cref{sec:split-prompt}), selects a non-final reasoning step based on its position-weighted uncertainty (\Cref{sec:uq}), proposes a deliberately distinct alternative candidate (\Cref{sec:diversity-prompt}), and then greedily decodes a new continuation. This procedure can be repeated for a fixed number of iterations or until a target number of accepted traces is obtained.
    }
    \label{fig:dart-overview}
\end{figure}

To further improve their reasoning capabilities, LLMs are commonly post-trained on reasoning traces \citep{lightmanLetsVerifyStep2023, wangMathShepherdVerifyReinforce2024, muennighoffS1SimpleTesttime2025}.
Since high-quality traces are generally difficult to obtain at scale, a common approach is to generate them from the model itself using Monte Carlo (MC) sampling at nonzero temperature.
The generated traces can then be evaluated using verifiable rewards, based on whether the final answer to a mathematical problem matches a known reference answer, and subsequently used for post-training \citep{zelikmanSTaRBootstrappingReasoning2022, deepseek-aiDeepSeekR1IncentivizingReasoning2025}.
We refer to traces as \textit{accepted} if they pass the verifier and \textit{rejected} otherwise.

A key limitation is that this approach provides no post-training data for problems where MC sampling fails to generate an accepted trace \citep{yuDAPOOpenSourceLLM2025, zhengActOnlyWhen2025, kirk2024understandingeffectsrlhfllm,cui2025entropy}, which is often the case for the hard mathematical problems we consider (\Cref{sec:mining}).
Moreover, repeated sampling may produce similar traces, leaving alternative reasoning strategies unexplored.
This turns trace generation into a question of problem coverage. A problem contributes post-training data only if the model generates at least one accepted trace. This creates a model-dependent data collection process that can underrepresent problems for which the model has a low probability of generating an accepted trace.

Let $q_{\boldsymbol{\theta}}(\mathbf{x})$ be the success probability of a fixed decoding policy on problem $\mathbf{x}$. With $k$ independent traces from MC sampling, the probability of obtaining at least one accepted trace is
\begin{equation}
1-\bigl(1-q_{\boldsymbol{\theta}}(\mathbf{x})\bigr)^k.
\label{eq:mc-coverage}
\end{equation}
When $k \, q_{\boldsymbol{\theta}}(\mathbf{x})\ll 1$, \Cref{eq:mc-coverage} is approximately $k \,q_{\boldsymbol{\theta}}(\mathbf{x})$, so MC sampling leaves many low-$q_{\boldsymbol{\theta}}$ problems without an accepted trace at practical $k$ \citep{yuDAPOOpenSourceLLM2025,zhengActOnlyWhen2025}.
Moreover, MC sampling starts each attempt from scratch, without leveraging the structure of previous failed attempts or controlling which local decision to change.

To address this, we argue that low-probability regions should be explored more deliberately by increasing diversity in the traces.
We propose \fullmethod~(\method), a trace-mining method for this setting. Starting from a rejected trace generated by an LLM, \method{} decomposes the trace into reasoning steps, ranks these steps by position-weighted uncertainty, and selects uncertain reasoning steps for branching. At a selected reasoning step, \method{} leverages the model's own reasoning capabilities to propose a conceptually distinct alternative candidate conditioned on the preceding steps, the selected step, and the reference answer, with a meta-thought guiding the alternative strategy. This prompts the model to depart from the original greedy trace. The model then greedily completes the trace from this alternative candidate onward (see \Cref{fig:dart-overview}).
We evaluate \method{} through component analyses on MATH-500 (\Cref{sec:method}), problem coverage and generation cost on two mining datasets (\Cref{sec:mining}), and subsequent post-training (\Cref{sec:post-training}).

Across multiple mathematical reasoning datasets, \method{} recovers problems missed by full-trace MC sampling. Ablations show that removing reference-answer guidance still yields improved problem coverage over matched step-level MC sampling. Supervised fine-tuning (SFT) on the resulting training corpora also yields higher downstream point estimates than MC sampling. Together, these findings support targeted exploration of rejected traces as a practical approach to collecting post-training data.

Our main findings are:
\begin{itemize}[leftmargin=12pt,topsep=0pt,itemsep=6pt,partopsep=0pt,parsep=0pt]
    \item At a matched number of completed traces, \method{} recovers more problems than MC sampling, with the largest gains on the hardest problems, reaching up to $2\times$ higher problem coverage (\Cref{fig:difficulty_small}).
    \item Position-weighted greedy negative log-likelihood (G-NLL$_{\texttt{POS}}$) reduces the mean number of alternative candidates required to obtain an accepted trace by about 40\% relative to random selection on problems with both accepted and rejected completed traces (\Cref{fig:uq}).
    \item \method{} recovers 37\% more problems and generates 45\% more accepted traces per recovered problem than MC sampling, yielding nearly twice as many accepted traces at a 30\% lower estimated generation cost per accepted trace, despite a 39\% higher cost per attempted problem (\Cref{fig:compute-ablation}).

    \item  In post-training evaluations, \method{} achieves up to 5 percentage points higher \emph{Greedy $\cup$ pass@32} accuracy than MC sampling and the highest point estimate in five of six evaluations (\Cref{tab:finetuning-main}).
\end{itemize}

\section{Method}
\label{sec:method}

Let the dataset $\mathcal{D}=\{(\mathbf{x}_n,\mathbf{a}_n)\}_{n=1}^N$ contain problems and reference answers, and let $p_{\boldsymbol{\theta}}$ denote a language model used to propose and complete traces.
For each problem $\mathbf{x}$, we first greedily generate a trace $\mathbf{t}$ using the chain-of-thought (CoT) instruction $\mathbf{p}_{\texttt{CoT}} = $
``\emph{Let's think step by step and output the final answer within \textbackslash boxed\{\}}''.
We then verify its final answer against the reference answer $\mathbf{a}$ using Math-Verify, denoted by $\smash{v(\mathbf{t},\mathbf{a})\in\{0,1\}}$\footnote{\url{https://github.com/huggingface/Math-Verify}}, treating traces that do not terminate normally as rejected.
We note that although this outcome verifier does not establish the validity of every intermediate reasoning step, it provides an automated signal for identifying traces that arrive at the correct final answer \citep{cobbeTrainingVerifiersSolve2021, lightmanLetsVerifyStep2023}.

An accepted greedy trace is retained directly; a rejected greedy trace is processed through reasoning step decomposition, reasoning step selection, and reasoning step diversity sampling (\Cref{fig:dart-overview}).

For a selected position $j$ and alternative candidate index $r$, the completed trace has the form
\begin{equation}
\mathbf{t}^{j,r}_{\texttt{ALT}}
=\mathbf{s}^{<j}\mathbin{\Vert}\mathbf{s}^{j,r}_{\texttt{ALT}}\mathbin{\Vert}\mathbf{c}^{j,r},
\label{eq:dart-candidate}
\end{equation}
where $\Vert$ denotes concatenation. The suffix $\mathbf{s}^{>j}$ of the rejected trace is discarded, allowing the alternative candidate to change subsequent reasoning rather than reconnecting to the original trace.

Unlike \emph{full-trace MC sampling}, which independently samples new traces from
$p_{\boldsymbol{\theta}}$, \method{} reuses the
prefix $\mathbf{s}^{<j}$ and generates a meta-thought, the alternative candidate
$\smash{\mathbf{s}^{j,r}_{\texttt{ALT}}}$, and the subsequent continuation
$\mathbf{c}^{j,r}$ with $p_{\boldsymbol{\theta}}$. This avoids regenerating
the shared prefix, although sampling $\smash{\mathbf{s}^{j,r}_{\texttt{ALT}}}$
incurs additional compute due to reasoning step diversity sampling. We analyze
this trade-off in \Cref{sec:mining}.

Consequently, the completed traces constructed by \method{} are not independent samples from $p_{\boldsymbol{\theta}}$ under the original decoding policy.
Instead, our method conditions on a previous rejected trace and targets uncertain reasoning steps at which an alternative candidate may lead to a different final answer. By deliberately departing from the original decoding policy, \method{} aims to improve problem coverage, particularly for problems with low $q_{\boldsymbol{\theta}}(\mathbf{x})$.
This locally induced shift in the sampling distribution is compatible with our post-training setup, where the training objective does not require on-policy sampling. We therefore retain every accepted trace for post-training.
We measure \emph{problem coverage} as the number of problems with at least one accepted trace. A problem is \emph{greedy-solved} if its greedy trace is accepted and \emph{greedy-unsolved} otherwise. \emph{Recovering} a greedy-unsolved problem means obtaining an accepted alternative trace.
We next detail the three stages used to construct these traces.

\subsection{Reasoning step decomposition} \label{sec:split-prompt}

To decompose a trace $\mathbf{t}$ into individual reasoning steps, one can use rule-based delimiters, such as line breaks \citep{fengAlphazerolikeTreeSearchCan2024, zhangReSTMCTSLLMSelfTraining2024, xieMonteCarloTree2024, tianSelfImprovementLLMsImagination2024}. However, such heuristics can fail to capture meaningful reasoning step boundaries. Multi-line sentences may be split into incomplete steps, while single lines containing multiple sentences may conflate distinct steps.

To address this, we use the language model itself to predict reasoning step boundaries. Specifically, we first split $\mathbf{t}$ deterministically at sentence-final punctuation and line breaks to obtain text fragments. We then use a reasoning step decomposition prompt $\mathbf{p}_{\texttt{AUG}}$ that guides the model to predict explicit \texttt{[STEP]} markers between text fragments. The model generates boundary decisions while the trace itself is teacher-forced. The model therefore classifies boundaries without regenerating or editing the trace.
Based on these markers, $\mathbf{t}$ is deterministically decomposed into reasoning steps $\smash{\{\mathbf{s}^j\}_{j=1}^{J(\mathbf{t})}}$, where $\smash{J(\mathbf{t})}$ is trace-dependent. This procedure preserves the original trace while allowing boundaries to depend on context (see \Cref{app:split-prompt} for $\mathbf{p}_{\texttt{AUG}}$ and examples).

Reasoning step decomposition adds only a small generation cost. The trace is teacher-forced, so each boundary decision decodes a single token, and the key-value (KV) cache of the annotated prefix can be reused across consecutive decisions. 
Decisions are batched across traces, only rejected greedy traces are decomposed, and decomposition during trace mining stops after ten closed reasoning steps (\Cref{app:split-prompt}). 
Since decomposition only places boundaries rather than generating reasoning, it need not use the policy model $p_{\boldsymbol{\theta}}$. A smaller model could further reduce this cost, but we leave this to future work.

\pagebreak
\subsection{Reasoning step selection} \label{sec:uq}

\begin{wrapfigure}{r}{0.5\textwidth}
    \centering
    \vspace{-13pt}
    \includegraphics[width=\linewidth,trim=0cm 0.42cm 0.36cm 0.4cm,clip]{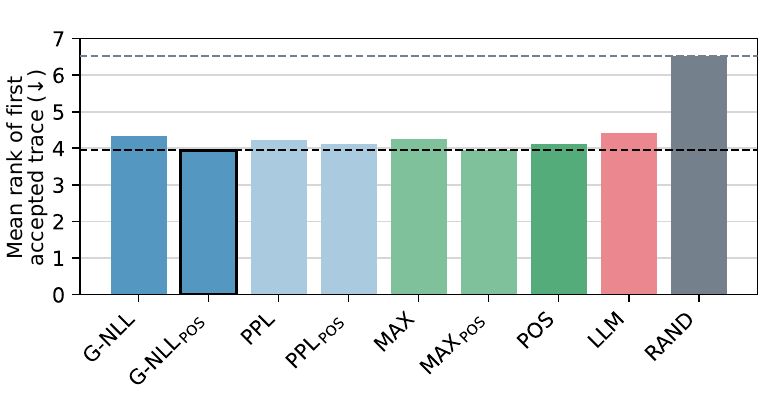}
    \caption{Mean rank of the first accepted trace on greedy-unsolved MATH-500 problems with both accepted and rejected completed traces. G-NLL, PPL, MAX: uncertainty; POS: position; LLM: self-reported confidence; RAND: random. Subscript POS denotes position weighting. Lower is better.}
    \label{fig:uq}
\end{wrapfigure}

Generating alternative candidates for all reasoning steps $\smash{\{\mathbf{s}^j\}_{j=1}^{J(\mathbf{t})}}$ would make the number of alternative candidates scale with trace length and become computationally expensive for long traces. We therefore select step positions $j$ that are most likely to yield a different reasoning strategy and, in turn, a different final answer. We consider several selection strategies guided by two main intuitions:\linebreak
(i) uncertain steps are promising to revisit, and (ii) alternative reasoning strategies are more likely to emerge earlier in the trace.

Among these strategies, the position-weighted G-NLL \citep{aichberger2024rethi} of each reasoning step (G-NLL$_{\texttt{POS}}$) performs best. Given $\mathbf{s}^j$, it is defined as
\begin{equation}
u(\mathbf{s}^{j}) =
- \ \frac{1}{j}
\sum_{i=1}^{|\mathbf{s}^{j}|}
\log p_{\boldsymbol{\theta}}
\left(s_i^{j}\mid \mathbf{x},\mathbf{s}^{<j},\mathbf{s}_{<i}^{j}\right).
\label{eq:step-uncertainty}
\end{equation}
We select steps with the largest scores. Dividing by $j$ favors early uncertain steps, where replacing the original reasoning step can redirect a larger portion of the remaining trace.
We compare rankings on greedy-unsolved MATH-500 problems with both accepted and rejected completed traces, the only problems on which the ranking can change the rank of the first accepted trace (125 problems; \Cref{fig:uq}, \Cref{app:step-selection}). On this subset, G-NLL$_{\texttt{POS}}$ places the first accepted trace at a mean rank slightly below four, compared with nearly seven under random selection, a reduction of about 40\%. This measures ranking effectiveness within the recorded sets of completed traces, rather than overall search cost across all greedy-unsolved problems. 
All uncertainty- and position-based scores yield lower mean ranks than random selection and perform similarly, indicating that the benefit of targeted selection is robust to the choice of score, while self-reported confidence performs slightly worse. We use the best-performing score, G-NLL$_{\texttt{POS}}$, throughout the remaining experiments.

\subsection{Reasoning step diversity sampling} \label{sec:diversity-prompt}

Once we have selected a reasoning step $\mathbf{s}^{j}$, we aim to generate an alternative candidate $\mathbf{s}^{j,r}_{\texttt{ALT}}$ that follows a different reasoning strategy. A simple baseline is \emph{step-level MC sampling}, which samples from $p_{\boldsymbol{\theta}}(\cdot\mid\mathbf{x},\mathbf{s}^{<j})$ at temperature $T$. While this introduces stochasticity, it does not condition on the original step $\mathbf{s}^{j}$ and therefore cannot explicitly encourage the alternative candidate to differ from it.

We thus introduce a diversity prompt $\mathbf{p}_{\texttt{DIV}}$ that includes $\mathbf{s}^{j}$ as contrastive context and instructs the model to construct an alternative candidate based on a fundamentally different reasoning strategy. The alternative candidate should form a single self-contained reasoning step, continue the retained prefix $\mathbf{s}^{<j}$ in the same format and style, replace $\mathbf{s}^{j}$ without repeating or referring back to it, and offer a non-obvious perspective that $\mathbf{s}^{j}$ does not anticipate (\Cref{app:diversity-sampling}). Alternative candidates are generated sequentially: for alternative candidate $r$, the diversity prompt includes the selected reasoning step and all earlier alternative candidates. This changes the context even under greedy decoding, encouraging distinct alternative candidates without guaranteeing semantic diversity.

\begin{wrapfigure}{r}{0.5\textwidth}
    \centering
    \includegraphics[width=\linewidth,trim=0cm -0.05cm 0.35cm 0.39cm,clip]{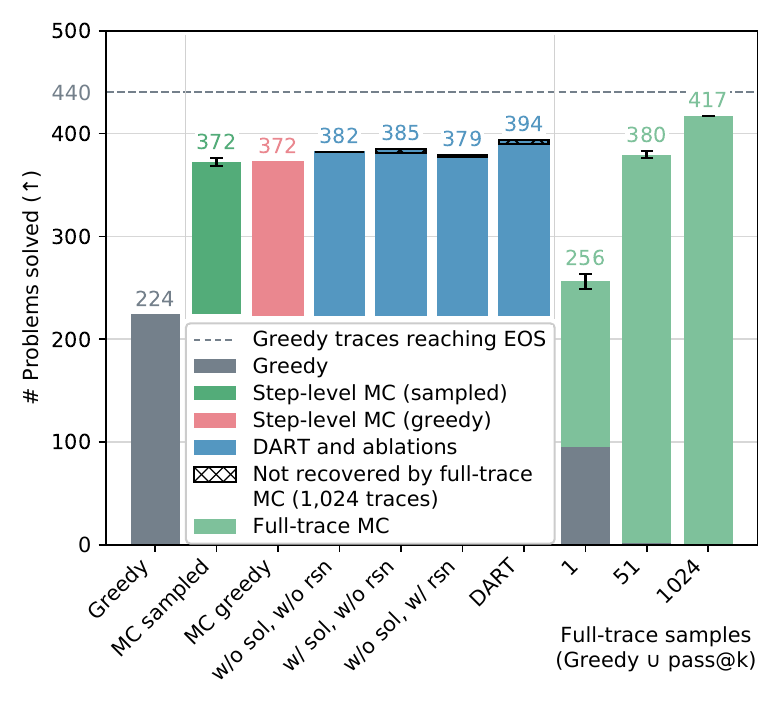}
    \caption{MATH-500 problem coverage, including 224 problems solved by greedy decoding (gray). We compare \method{} and its ablations (blue) against step-level MC sampling with sampled or greedy continuation (dark green or red) and full-trace MC sampling (light green). The full-trace axis reports Greedy $\cup$ pass@$k$. For the ablations, \texttt{sol} denotes reference-answer guidance, while \texttt{rsn} denotes generating a meta-thought to identify promising alternative candidates. Hatching indicates problems recovered by a method but not by full-trace MC sampling with 1,024 traces.}
    \label{fig:ablation-math-500}
    \vspace{-8pt}
\end{wrapfigure}

In \method{} (\texttt{w/ sol, w/ rsn}), the prompt additionally provides the reference answer $\mathbf{a}$, with instructions not to reveal it, and asks the model to first generate a meta-thought about a suitable alternative strategy (\emph{Meta-Thought} in \Cref{fig:dart-overview}). Both components only shape the proposal. The meta-thought is discarded, and the continuation $\mathbf{c}^{j,r}$ is greedily generated from $\mathbf{s}^{<j}\mathbin{\Vert}\mathbf{s}^{j,r}_{\texttt{ALT}}$ under the original instruction $\mathbf{p}_{\texttt{CoT}}$, without access to $\mathbf{p}_{\texttt{DIV}}$, $\mathbf{a}$, or the meta-thought. Reference-answer guidance therefore enters the completed trace only through the alternative candidate. We ablate these two components in a $2\times2$ comparison, removing reference-answer guidance, the meta-thought, or both (\Cref{fig:ablation-math-500}).

\pagebreak

The diversity prompt $\mathbf{p}_{\texttt{DIV}}$ encourages a qualitative change in reasoning strategy rather than optimizing an explicit semantic-distance objective. We therefore assess its effectiveness through problem coverage. \Cref{fig:ablation-math-500} compares \method{} and its three ablations (blue) on MATH-500 using Llama-3.1-8B-Instruct \citep{grattafiori2024,lightmanLetsVerifyStep2023}. Our primary baseline, step-level MC sampling with greedy continuation (red), samples an alternative candidate at $T=1$ and continues greedily like \method{}, so that the comparison isolates the proposal procedure at a matched completed-trace budget. Removing both reference-answer guidance and the meta-thought gives the closest ablation to this baseline. We also compare against step-level MC sampling with sampled continuation (dark green), which samples both at $T=1$, and full-trace MC sampling (light green), which generates each trace from scratch.

Among problems with normally terminated greedy traces, \method{} recovers 170 problems at the largest evaluated step-level budget, compared with 148 for step-level MC sampling with greedy continuation, about 15\% more recoveries. 
The ablations without reference-answer guidance recover 158 problems without a meta-thought and 155 with one, showing that reference-answer guidance helps but is not necessary for improved recovery.

With one to four alternative candidates per selected reasoning step, \method{} covers 350, 373, 386, and 394 problems (including 224 greedy-solved problems), versus 332, 356, 365, and 372 for step-level MC sampling with greedy continuation (\Cref{fig:app_ablation-math-500}). At the largest budget, its problem coverage also exceeds that of full-trace MC sampling with 51 traces. Some \method{} recoveries are missed even by full-trace MC sampling with 1,024 traces, illustrating complementary problem coverage (\Cref{app:diversity-ablation}).
\Cref{alg:dart} brings the three stages together into the complete trace-mining procedure.

\begin{algorithm}[ht]
\caption{\method{} trace mining for one problem}
\label{alg:dart}
\begin{algorithmic}[1]
\Require Problem $\mathbf{x}$, reference answer $\mathbf{a}$, model $p_{\boldsymbol{\theta}}$, step budget $m$, and $R$ alternative candidates  %
\Ensure Set $\mathcal{T}$ of accepted traces, each paired with $\mathbf{x}$ and $\mathbf{p}_{\texttt{CoT}}$ for SFT
\State $\mathbf{t} \gets$ greedy trace generated with $\mathbf{p}_{\texttt{CoT}}$
\IfThen{$v(\mathbf{t},\mathbf{a})=1$}{\Return $\{\mathbf{t}\}$}
\State $\mathcal{T} \gets \emptyset$
\State $\{\mathbf{s}^{j}\}_{j=1}^{J(\mathbf{t})} \gets$ decompose $\mathbf{t}$ into reasoning steps \Comment{\Cref{sec:split-prompt}}
\State $\mathcal{J} \gets$ positions of up to $m$ non-final reasoning steps with the largest $u(\mathbf{s}^{j})$ \Comment{\Cref{eq:step-uncertainty}}
\For{$j \in \mathcal{J}$}
    \State $\mathcal{H} \gets \{\mathbf{s}^{j}\}$ \Comment{candidate history}
    \For{$r = 1, \ldots, R$}
        \State generate a meta-thought and $\mathbf{s}^{j,r}_{\texttt{ALT}}$ with $\mathbf{p}_{\texttt{DIV}}(\mathbf{x}, \mathbf{a}, \mathbf{s}^{<j}, \mathcal{H})$ at $T=0$ \Comment{\Cref{sec:diversity-prompt}}
        \State $\mathcal{H} \gets \mathcal{H} \cup \{\mathbf{s}^{j,r}_{\texttt{ALT}}\}$
        \State $\mathbf{c}^{j,r} \gets$ greedy continuation of $\mathbf{s}^{<j} \mathbin{\Vert} \mathbf{s}^{j,r}_{\texttt{ALT}}$ under $\mathbf{p}_{\texttt{CoT}}$
        \State $\mathbf{t}^{j,r}_{\texttt{ALT}} \gets \mathbf{s}^{<j} \mathbin{\Vert} \mathbf{s}^{j,r}_{\texttt{ALT}} \mathbin{\Vert} \mathbf{c}^{j,r}$ \Comment{\Cref{eq:dart-candidate}}
        \IfThen{$v(\mathbf{t}^{j,r}_{\texttt{ALT}}, \mathbf{a}) = 1$}{$\mathcal{T} \gets \mathcal{T} \cup \{\mathbf{t}^{j,r}_{\texttt{ALT}}\}$}
    \EndFor
\EndFor
\State \Return $\mathcal{T}$
\end{algorithmic}
\end{algorithm}

\section{Related work}

\paragraph{Diverse reasoning and verified post-training.}
Chain-of-thought prompting elicits intermediate reasoning \citep{weiChainofThoughtPromptingElicits2023}, and self-consistency samples multiple traces and aggregates their answers \citep{wangSelfConsistencyImprovesChain2023}. Other methods diversify decompositions, demonstrations, or solution paths at inference or during training \citep{zhou2022least,press2022measuring,zhang2022automatic,liMakingLanguageModels2023,liMuggleMathAssessing2024,yuFlowReasoning2025,dokmeciDataDiversification2025}. Rejection fine-tuning and ReST retain accepted model-generated traces for subsequent learning \citep{yuanRFT2023,gulcehreReinforcedSelfTrainingReST2023}. STaR additionally regenerates rejected traces conditioned on the reference answer, then fine-tunes on accepted traces and repeats generation, combining reference-answer guidance during data acquisition with iterative model improvement \citep{zelikmanSTaRBootstrappingReasoning2022}. Related work improves trace selection with verifiers or process supervision \citep{hosseiniVSTaRTrainingVerifiers2024,lightmanLetsVerifyStep2023,wangMathShepherdVerifyReinforce2024,luoAutomatedProcessSupervision2024}, or optimizes directly against verifiable rewards \citep{shaoDeepSeekMathPushingLimits2024,deepseek-aiDeepSeekR1IncentivizingReasoning2025,yuDAPOOpenSourceLLM2025}. \method{} targets the preceding data-acquisition step. It shares STaR's reference-answer guidance, but proposes a local alternative candidate from a rejected trace rather than regenerating the full trace. The resulting accepted traces can feed standard supervised or reinforcement-learning objectives.

\vspace{-1pt}
\paragraph{Guided search.}
Tree-of-Thoughts and RAP organize reasoning as search over intermediate states \citep{yaoTreeThoughtsDeliberate2023,haoReasoningLanguageModel2023}. Subsequent systems combine tree search with learned critics, process rewards, self-training, or preference optimization \citep{fengAlphazerolikeTreeSearchCan2024,tianSelfImprovementLLMsImagination2024,xieMonteCarloTree2024,yangTreeRPO2025,guanRStarMath2025}. For example, ReST-MCTS* alternates process-reward-guided search with training of both the policy and process reward model \citep{zhangReSTMCTSLLMSelfTraining2024}. Contract-checked graph editing instead recombines partial traces under structural and deterministic checks \citep{liContractCheckedEditing2026}. Uncertainty-guided methods use local uncertainty to decide where to branch \citep{moTreeUncertainThoughts2023,liEntropyAwareBranching2025}. Entropy-Gated Branching branches only at high-uncertainty steps and prunes expansions with a lightweight verifier, while ROSE uses semantic entropy to allocate branches within reinforcement learning \citep{liEntropyAwareBranching2025,zhaoReinforcedEfficientReasoning2026}. \method{} builds on this line of work but addresses a complementary question: how to propose an alternative candidate once a reasoning step is selected. The selected reasoning step serves as contrastive context for the proposal, while acceptance depends only on the completed trace's final answer. This requires neither learned step rewards nor critic updates. Our step-level comparisons fix the model, verifier, problems, and completed-trace budget. They therefore study the proposal mechanism rather than compare complete search systems that also introduce critics, backups, or new training objectives.

\vspace{-1pt}
\paragraph{Reflection and correction.}
Self-Refine uses the same model to generate feedback and iteratively revise a response \citep{madaanSelfRefineIterativeRefinement2023}. Reflexion stores linguistic feedback across attempts \citep{shinnReflexionLanguageAgents2023}. Other methods learn to critique, correct, or check generated responses \citep{saundersSelfcritiquingModelsAssisting2022,welleckGeneratingSequencesLearning2022,zhangLearningCheckUnleashing2024,wangCritiqueFineTuningLearning2025,liReflectEvoImprovingMeta2025}. \method{} instead retains the prefix before a selected reasoning step, replaces that reasoning step, and discards the original suffix. Its meta-thought guides this local proposal without asserting that the selected reasoning step is the first error. Only the resulting accepted trace, not the meta-thought or rejected trace, is used for SFT.

\section{Mining traces with \method}
\label{sec:mining}

The MATH-500 component analyses examine where to branch and how to construct alternative candidates. We now evaluate whether the complete \method{} procedure expands problem coverage on larger mining datasets and how this affects generation cost. We compare it with full-trace MC sampling, hereafter MC sampling, on the mining datasets DeepMath-103K \citep{he2025deepmath} and OpenMathInstruct-2 \citep{toshniwal2024openmathinstruct}, using Llama-3.1-8B-Instruct to make both the 512-trace difficulty analysis and post-training tractable.

\begin{figure*}[tb]
    \centering
    {\small
    \begin{tabular*}{\textwidth}{@{\extracolsep{\fill}}lll@{}}
        \colorswatch{AppleBlue5}~~\method{}, first selected reasoning step &
        \colorswatch{AppleBlue3}~~\method{}, second selected reasoning step &
        \colorswatch{AppleGreen4}~~both methods \\
        \colorswatch{AppleRed4}~~MC sampling only &
        \crossedswatch{AppleRed4}~~\method{} only
    \end{tabular*}
    }\\
    \vspace{4pt}
    \begin{subfigure}{0.5\textwidth}
        \includegraphics[trim={0.8cm 0 0 4.3cm},clip,width=\columnwidth]{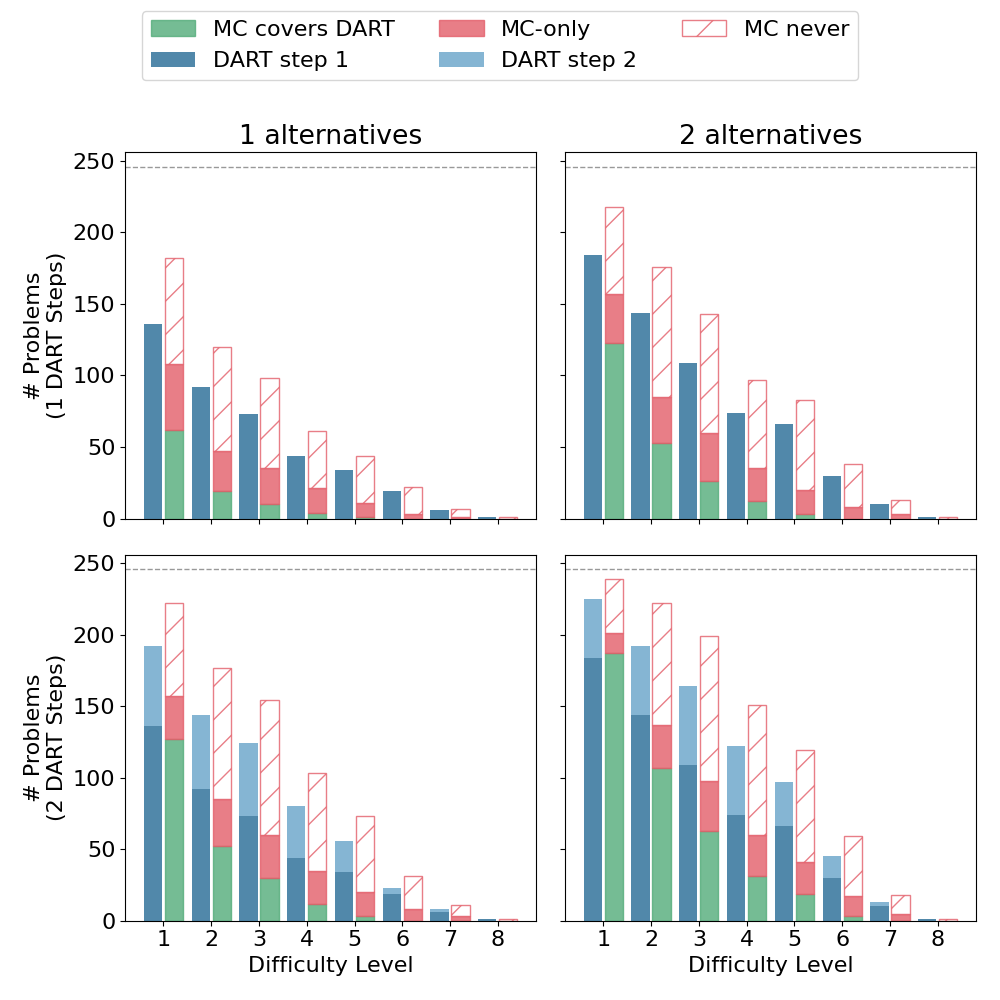}
        \caption{DeepMath-103K \citep{he2025deepmath}}
    \end{subfigure}\hfill
    \begin{subfigure}{0.5\textwidth}
        \includegraphics[trim={0cm 0 0.8cm 4.3cm},clip,width=\columnwidth]{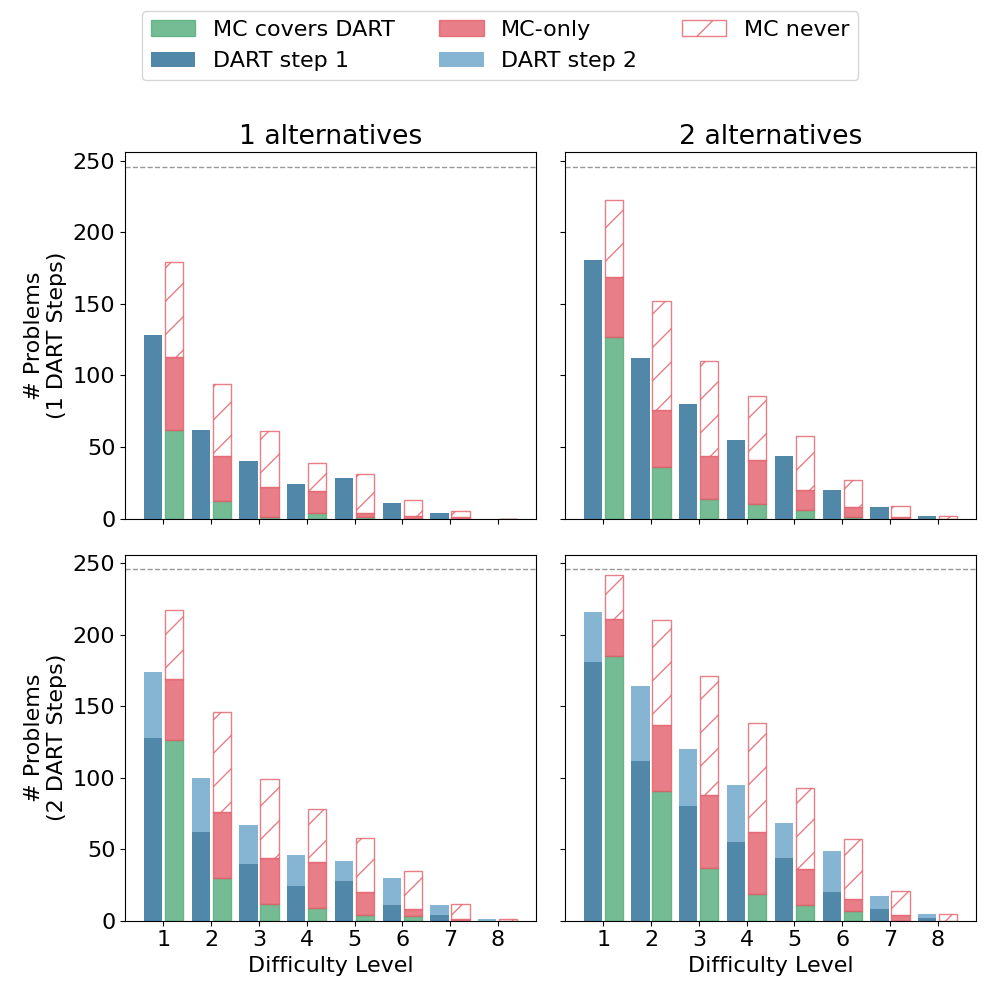}
        \caption{OpenMathInstruct-2 \citep{toshniwal2024openmathinstruct}}
    \end{subfigure}
    \caption{Number of problems recovered by \method{} and MC sampling at matched numbers of completed traces. All problems are greedy-unsolved. Blue: \method{} recoveries by the first (dark) and second (light) selected reasoning step; green: recovered by both methods; red: recovered by MC sampling only; red hatching: recovered by \method{} only. %
    Difficulty strata and MC sampling results are based on the same 512-trace pool, so MC sampling has zero recoveries by construction in the hardest stratum.}
    \label{fig:difficulty_small}
    \vspace{-2pt}
\end{figure*}

\vspace{-1pt}
\paragraph{Evaluation unit and metric.}
Throughout this section, we evaluate only greedy-unsolved problems. At a budget of $k$ completed traces per problem, problem coverage is the number of recovered problems, each with at least one accepted trace. For MC sampling before stratification, the per-problem recovery probability is given by \Cref{eq:mc-coverage}. Problem coverage counts a recovered problem once, irrespective of how many accepted traces it yields, and therefore does not measure semantic diversity among traces. It thus measures how many previously uncovered problems contribute post-training data.

\paragraph{Model-adjusted difficulty.}
To localize the gains, we stratify problems by $q_{\boldsymbol{\theta}}(\mathbf{x})$ from \Cref{sec:introduction}. Human-oriented labels need not align with this model-dependent quantity (\Cref{fig:deepmath_original_difficulty} in \Cref{app:inference}). For each pair $(\mathbf{x},\mathbf{a})$, we estimate $q_{\boldsymbol{\theta}}(\mathbf{x})$ from a pool of 512 independent traces generated by MC sampling at $T=1$,
\begin{equation}
    \widehat q(\mathbf{x}) = \frac{1}{512}\sum_{\ell=1}^{512} v(\mathbf{t}^{\ell},\mathbf{a}),
    \qquad \mathbf{t}^{\ell} \sim p_{\boldsymbol{\theta}}(\cdot\mid \mathbf{x};T=1).
    \label{eq:model-adjusted-difficulty}
\end{equation}
Using the fixed ranges in \Cref{app:inference}, we form eight strata of up to 255 greedy-unsolved problems per dataset, ordered from easiest to hardest. This subsampling does not affect the generated traces or the training corpora (\Cref{sec:post-training}). The displayed results of MC sampling use traces from this same 512-trace pool, so stratification and evaluation are not independent. In particular, by construction, MC sampling recovers no problem with $\widehat q(\mathbf{x})=0$. The strata describe empirical sampling difficulty, so zero observed successes does not imply zero underlying success probability.

\vspace{-4pt}
\paragraph{Matched completed-trace budget.}
MC sampling draws $k\in\{1,2,4\}$ independent full traces per problem at $T=1$. \method{} uses the same completed-trace budget $k$, distributing it across selected reasoning steps and alternative candidates. At $k=2$, we evaluate both two alternative candidates for one selected reasoning step and one alternative candidate for each of two selected reasoning steps, separating the benefit of exploring multiple alternative candidates at a single step from intervening at multiple steps. This design allows each \method{} configuration to be compared directly with MC sampling at the same number of completed traces.

\vspace{-4pt}
\paragraph{Results.}
\Cref{fig:difficulty_small} shows that \method{} consistently achieves higher aggregate problem coverage than MC sampling across all displayed budget allocations on both datasets, with results for larger budgets provided in \Cref{fig:open_math_difficulty,fig:deep_math_difficulty} (\Cref{app:inference}).
Increasing the completed-trace budget further improves coverage: both $k=2$ allocations outperform $k=1$, while generating two alternative candidates at each of two selected reasoning steps ($k=4$) achieves the highest coverage.
These results support exploring multiple alternative candidates both within and across reasoning steps.

Notably, \method{} also recovers problems in the hardest strata. The methods recover partially overlapping sets of problems, with some recovered exclusively by \method{} and others exclusively by MC sampling.
Together, these findings highlight that targeted generation of alternative candidates recovers problems that are difficult to reach through independent sampling, while suggesting that the two approaches could be combined to further expand problem coverage.

\begin{figure*}[t]
    \centering
    \captionsetup[subfigure]{justification=centering}
    \begin{subfigure}[t]{0.25\textwidth}
        \centering
        \includegraphics[width=\linewidth]{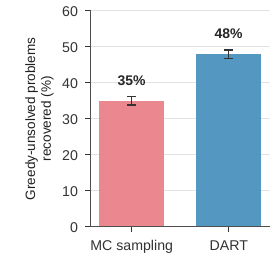}
        \caption{Problem recovery}
        \label{fig:compute-recovery}
    \end{subfigure}\hfill
    \begin{subfigure}[t]{0.25\textwidth}
        \centering
        \includegraphics[width=\linewidth]{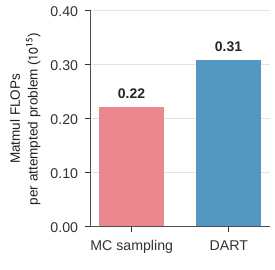}
        \caption{Cost per attempted problem}
        \label{fig:compute-per-problem}
    \end{subfigure}\hfill
    \begin{subfigure}[t]{0.25\textwidth}
        \centering
        \includegraphics[width=\linewidth]{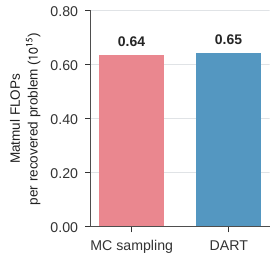}
        \caption{Cost per recovered problem}
        \label{fig:compute-per-recovery}
    \end{subfigure}\hfill
    \begin{subfigure}[t]{0.25\textwidth}
        \centering
        \includegraphics[width=\linewidth]{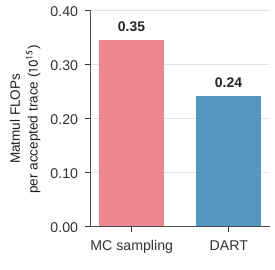}
        \caption{Cost per accepted trace}
        \label{fig:compute-per-trace}
    \end{subfigure}
    \vspace{2pt}
    \caption{Coverage and estimated generation cost on 6,200 shared greedy-unsolved DeepMath-103K problems, using approximately eight completed traces per problem. Blue denotes \method{}; red denotes MC sampling. (\subref{fig:compute-recovery}) Fraction recovered by at least one accepted trace. Error bars are 95\% Wilson intervals. (\subref{fig:compute-per-problem}--\subref{fig:compute-per-trace}) Estimated matrix-multiplication FLOPs per attempted problem, recovered problem, and accepted trace, respectively. Panel (\subref{fig:compute-per-trace}) counts multiple accepted traces from one problem, without semantic deduplication.}
    \label{fig:compute-ablation}
\end{figure*}

\paragraph{Generation-cost analysis.}
The primary comparison controls for the number of completed traces but not differences in computational cost. We therefore analyze generation cost (\Cref{fig:compute-ablation}) on greedy-unsolved problems from DeepMath-103K.
MC sampling generates eight full traces per problem at $T=1$. With a budget of up to eight completed traces, \method{} selects up to four reasoning steps and generates two alternative candidates per step. For each completed trace, the model first generates a meta-thought and alternative candidate at $T=0$, followed by a greedy continuation. This yields up to 49,600 completed traces for \method{} and MC sampling. Both methods use the same 4,096-token completion allowance. For \method{}, retained prefixes and alternative candidates count toward this allowance, with retained-prefix KV caches being reused to avoid redundant computation.

We estimate generation FLOPs accounting for differences in sequence length and prefix reuse. The estimate includes alternative candidate generation and trace completion but excludes reasoning step decomposition, the shared initial greedy attempt, and non-generation overhead. It therefore does not measure end-to-end runtime (see \Cref{app:compute-accounting} for details).

\paragraph{Coverage and cost.}
\Cref{fig:compute-ablation} shows that \method{} recovers 2,971 of 6,200 problems (48\%), compared with 2,168 (35\%) for MC sampling, corresponding to 37\% more recovered problems. This increased coverage comes at a 39\% higher estimated generation cost per attempted problem. However, the cost per recovered problem remains nearly unchanged, while the cost per accepted trace decreases by 30\% because \method{} produces almost twice as many accepted traces.
These results demonstrate that targeted exploration can substantially improve both problem recovery and accepted-trace yield without a proportional increase in estimated generation cost. In particular, the additional computation required to construct alternative candidates is effectively offset by improved problem coverage, allowing \method{} to recover substantially more problems at nearly the same estimated generation cost per recovered problem.

\paragraph{Explaining the cost trade-off.}
The higher cost per attempted problem can be attributed primarily to the number of processed tokens. Projection and MLP operations account for approximately 88\% and 91\% of the estimated FLOPs for MC sampling and \method{}, respectively. \method{} processes 43\% more uncached prompt and generated tokens than MC sampling, as each completed trace involves a diversity prompt and a meta-thought. However, it requires only 9\% more causal token pairs, since reusing retained-prefix KV caches avoids recomputing attention over the shared prefix (\Cref{tab:compute-counters}).

The higher accepted-trace yield reflects two effects. \method{} recovers $1.37\times$ as many problems as MC sampling and yields $2.67$ accepted traces per recovered problem, compared with $1.84$ for MC sampling (\Cref{tab:compute-counters}), a further $1.45\times$ increase. Together, these factors explain the nearly twofold increase in accepted traces.
Although the accepted traces are not semantically deduplicated, \method{} provides both broader problem coverage and more accepted reasoning traces per recovered problem for subsequent post-training.

\section{Post-training on mined \method{} traces}
\label{sec:post-training}

We next evaluate the downstream value of traces mined by \method{} through supervised fine-tuning (SFT) of Llama-3.1-8B-Instruct. We mine separate training corpora from 8k problems drawn from DeepMath-103K \citep{he2025deepmath} and 16k problems drawn from OpenMathInstruct-2 \citep{toshniwal2024openmathinstruct}. Within each mining dataset, SFT on the training corpus obtained from each of the following mining methods starts from the same base checkpoint and uses the same SFT recipe and random seed:

\begin{itemize}[leftmargin=12pt,topsep=0pt,itemsep=6pt,partopsep=0pt,parsep=0pt]
    \item \textbf{Greedy:} All accepted greedy traces. This baseline training corpus is included in every other training corpus.
    \item \textbf{MC sampling:} Eight independent full traces at $T=1$ for each rejected greedy trace, representing standard rejection sampling through repeated stochastic generation.
    \item \textbf{STaR:} One rationalization with reference-answer guidance for each rejected greedy trace, using the reference answer to elicit an accepted trace \citep{zelikmanSTaRBootstrappingReasoning2022}.
    \item \textbf{RFT:} Eight full traces at $T=1$ with reference-answer guidance for each rejected greedy trace, combining rejection sampling with access to the reference answer \citep{yuanRFT2023}. %
    \item \textbf{\method:} Two alternative candidates at each of up to four selected reasoning steps, followed by greedy continuation, yielding up to eight completed traces per rejected greedy trace.
\end{itemize}

STaR and RFT denote adapted mining baselines. Both receive the reference answer, as does \method{}. For \method{}, the meta-thought guides the proposal of an alternative candidate but is excluded from the training corpus. Each mining method retains only accepted traces. The resulting training corpora are not matched in examples or tokens because the mining methods recover different problems and yield different numbers of accepted traces. This end-to-end comparison evaluates the practical value of each mining method through downstream performance after SFT on its training corpus. 
With a fixed two-epoch SFT recipe, this comparison captures the combined benefits of the generated training corpora in terms of problem coverage, corpus size, and corpus composition, including the resulting differences in training volume.

\begin{table*}[t!]
    \centering
    \caption{Greedy $\cup$ pass@32 accuracy (\%) after SFT of Llama-3.1-8B-Instruct on traces mined from 16k OpenMathInstruct-2 or 8k DeepMath-103K problems. Base denotes the model before SFT. Each entry is the highest point estimate across evaluated checkpoints; bold and underline mark the two largest per column. Brackets give central 95\% ranges over repeated draws of 32 stochastic traces without replacement from the recorded pools, with a separate accepted greedy trace as an override. They exclude training-seed and checkpoint-selection uncertainty.}
    \label{tab:finetuning-main}
    \renewcommand{\arraystretch}{1.3}
    \setlength{\tabcolsep}{1pt}
    \begin{tabular}{lcccccc}
    \toprule
         &  \multicolumn{3}{c}{OpenMathInstruct-2 \citep{toshniwal2024openmathinstruct}} & \multicolumn{3}{c}{DeepMath-103K \citep{he2025deepmath}}\\
         \cmidrule(lr{4pt}){2-4} \cmidrule(lr{4pt}){5-7}
        Method & MATH-500 & AIME 22+23 & AIME 24+25 & MATH-500 & AIME 22+23 & AIME 24+25\\
       \midrule
        Base & 82.23\,{\scriptsize[82--85]} & 15.90\,{\scriptsize[12--20]} & 16.94\,{\scriptsize[12--22]} & 82.23\,{\scriptsize[82--85]} & 15.90\,{\scriptsize[12--20]} & 16.94\,{\scriptsize[12--22]}\\
        Greedy & 87.73\,{\scriptsize[86--89]} & 19.61\,{\scriptsize[15--23]} & 23.52\,{\scriptsize[17--30]} & 87.18\,{\scriptsize[86--88]} & 22.60\,{\scriptsize[17--28]} & 22.62\,{\scriptsize[18--28]}\\
        MC sampling & 87.09\,{\scriptsize[86--88]} & \underline{21.66}\,{\scriptsize[17--27]} & 23.05\,{\scriptsize[18--28]} & \underline{87.47}\,{\scriptsize[86--89]} & 22.88\,{\scriptsize[18--27]} & \underline{23.02}\,{\scriptsize[18--27]}\\
        STaR & \textbf{88.12}\,{\scriptsize[87--89]} & 19.95\,{\scriptsize[15--25]} & \underline{26.45}\,{\scriptsize[20--32]} & 86.92\,{\scriptsize[86--88]} & 20.80\,{\scriptsize[15--25]} & 22.22\,{\scriptsize[17--27]}\\
        RFT & 84.21\,{\scriptsize[83--85]} & 14.13\,{\scriptsize[10--17]} & 19.07\,{\scriptsize[15--23]} & 84.66\,{\scriptsize[83--86]} & \underline{23.91}\,{\scriptsize[18--28]} & 18.85\,{\scriptsize[13--25]}\\
        \method{} & \underline{87.96}\,{\scriptsize[87--89]} & \textbf{24.25}\,{\scriptsize[18--30]} & \textbf{28.12}\,{\scriptsize[23--33]} & \textbf{87.92}\,{\scriptsize[87--89]} & \textbf{24.89}\,{\scriptsize[20--30]} & \textbf{23.73}\,{\scriptsize[18--28]}\\
       \bottomrule
    \end{tabular}
\end{table*}

\paragraph{Training and evaluation prompts.}
For SFT, each accepted trace is paired only with the original problem $\mathbf{x}$ and CoT instruction $\mathbf{p}_{\texttt{CoT}}$, excluding all mining-specific guidance. Evaluation uses the same task prompt and likewise provides neither the reference answer $\mathbf{a}$ nor the rejected trace, alternative candidate, or meta-thought. The reference answer $a$ is used only for verification.
Thus, the additional information available during mining serves exclusively to construct the training corpora and is unavailable to the fine-tuned model at inference time. The evaluation therefore measures generation under the original task setting.

\paragraph{Evaluation.}
We report Greedy $\cup$ pass@32 accuracy (\Cref{tab:finetuning-main}), counting a problem as solved if either a separate greedy trace or at least one of 32 stochastic traces is accepted. This metric captures both the model's default greedy behavior and accepted traces obtainable through sampling, making it particularly relevant when greedy decoding alone may not yield an accepted trace. All models are evaluated using the same decoding, verification, and trace-resampling procedures to ensure consistent comparisons. We report the best-performing checkpoint for each model on each evaluation set. The uncertainty intervals reflect only resampling of the recorded trace pools, excluding training-seed variability and checkpoint-selection uncertainty. See \Cref{app:posttraining-protocol} for the full training and evaluation protocol.

\vspace{-4pt}
\paragraph{Results.}
Across MATH-500 \citep{lightmanLetsVerifyStep2023} and AIME \citep{maaAIME}, \method{} achieves a higher point estimate than MC sampling in all six combinations of mining dataset and evaluation set (\Cref{tab:finetuning-main}). Overall, \method{} ranks first in five of six evaluations, including all four AIME evaluations, with STaR only slightly outperforming it on MATH-500 after OpenMathInstruct-2 training. MATH-500 was used to develop the diversity prompt and reasoning step selection rule, while AIME informed neither choice. The AIME results therefore extend the evaluation beyond the benchmark used for method development, although AIME scores were used for checkpoint selection. \Cref{fig:best_pass_at_k} in \Cref{app:posttraining-protocol} additionally reports Greedy $\cup$ pass@$k$ curves, including cross-set evaluation of checkpoints selected on a single evaluation set.

STaR and RFT provide baselines with reference-answer guidance during mining. \method{} exceeds RFT in all six settings and STaR in five, showing favorable downstream point estimates against these baselines as well. Together, these results support targeted trace mining as an effective way to obtain post-training data.

\vspace{-1pt}
\section{Limitations}
\label{sec:limitations}
\vspace{-1pt}

\method{} uses the reference answer when constructing alternative candidates, making it primarily suited to trace mining for post-training rather than test-time reasoning. However, ablations without reference-answer guidance still outperform matched step-level MC sampling (\Cref{fig:ablation-math-500}). We verify final-answer correctness and evaluate exploration through problem coverage, which directly aligns with our objective of mining accepted traces. While these measures do not guarantee the validity of every intermediate reasoning step, this limitation is common to outcome-based verification approaches.

Our main coverage comparison controls for the number of completed traces, while a dedicated generation-cost analysis separately characterizes the compute trade-off at one representative operating point. The FLOP estimate excludes reasoning step decomposition and does not measure runtime. The post-training comparison evaluates complete mining methods, where corpus size, composition, and training volume vary together. This reflects the intended use of \method{}, where both broader problem coverage and a higher yield of accepted traces contribute to the value of the mined training corpus.

The empirical evaluation focuses primarily on one model and mathematical reasoning. While MATH-500 informed method development, the results on AIME extend the evaluation beyond this benchmark. Post-training uses one seed and evaluation-set checkpoint selection, and exact-string deduplication cannot exclude near-duplicate contamination. This motivates further investigation of \method{} across model scales and domains. Future work could also explore adaptive trace mining throughout post-training, dynamically targeting problems that remain challenging as model capabilities evolve.

\vspace{-1pt}
\section{Conclusion}
\label{sec:conclusion}
\vspace{-1pt}

We introduced \method{}, a trace-mining method that revisits uncertain reasoning steps in rejected traces and uses reference-answer guidance and a meta-thought to construct alternative candidates. \method{} improves MATH-500 recovery over matched step-level MC sampling, with ablations showing gains even without reference-answer guidance. Across the studied mathematical datasets, \method{} recovers problems missed by full-trace MC sampling, without a learned critic or process reward model.

On the 6,200 greedy-unsolved DeepMath-103K problems in the generation-cost analysis, \method{} recovers 37\% more problems at nearly the same estimated generation cost per recovered problem and 30\% lower cost per accepted trace. SFT on the resulting training corpora yields higher point estimates than MC sampling in all six evaluations and ranks first in five, supporting the practical value of \method{} for mining post-training data. These findings establish targeted exploration of rejected traces as a promising way to expand the supply of accepted traces for post-training.

\clearpage

\begin{ack}
This work was funded by Apple Inc. Lukas Aichberger and Jacob Springer conducted this work during internships at Apple. The authors declare no other competing interests.
\end{ack}

\bibliographystyle{main}
\bibliography{custom}

\clearpage
\appendix

\section{Reasoning step decomposition} \label{app:split-prompt}

Given a trace $\mathbf{t}$, the decomposition procedure is:
\begin{enumerate}[leftmargin=16pt,label=(\roman*),topsep=2pt,itemsep=1pt]
    \item Scan left to right and form text fragments at \texttt{.}, \texttt{!}, or \texttt{?} followed by whitespace or the end of the trace, and at line breaks. Delimiters and whitespace remain attached to the original text.
    \item Initialize the annotation with \texttt{[STEP]}. After appending each text fragment to the annotated prefix, greedily decode a single token from the decomposition prompt below.
    \item If the decoded token opens a \texttt{[STEP]} marker, close the accumulated reasoning step and insert the marker into the annotated prefix; otherwise append the next text fragment to the same reasoning step. Stop requesting boundary decisions after the configured number of closed reasoning steps or at the end of the trace, and retain any remaining text as the final reasoning step.
\end{enumerate}
The decoded token is used only as a boundary decision. Each reasoning step $\mathbf{s}^j$ is assembled from substrings of the input, so concatenation reconstructs $\mathbf{t}$ exactly. The implementation batches decisions at the same fragment index across traces. Post-training runs allow ten closed reasoning steps plus any remaining suffix; the final reasoning step is not eligible for selection. The annotation markers are not included in the training corpora.

\subsection{Prompt}

The decomposition prompt $\mathbf{p}_{\texttt{AUG}}$ is combined with the fixed demonstration below and the trace to annotate. The incremental procedure exposes only rule-based fragment boundaries and retains the original text regardless of the boundary decision.

\begin{promptbox}
\begin{prompttext}
You are an expert in analyzing reasoning steps. 
Your task is to annotate a step-by-step solution by inserting [STEP] markers to separate distinct reasoning steps.

RULES:
- Insert [STEP] at the beginning of each new reasoning step.
- A reasoning step is a single, self-contained logical unit such as:
  * Making an assumption or stating a premise
  * Reformulating or specifying the problem or previous steps
  * Performing a calculation or algebraic manipulation
  * Applying a rule, theorem, or logical principle
  * Executing a code block or multiple lines of code
  * Drawing a conclusion from previous steps
- Code: An entire code block or all subsequent lines of code must form one single, self-contained reasoning step.
- Text: Each sentence usually constitutes one single reasoning step. A reasoning step may only span multiple sentences if they form a single, inseparable operation or thought.
- Whenever it is ambiguous whether a new reasoning step begins, always prefer inserting a [STEP] marker.
- Do not edit, paraphrase, reorder, or replace any original text. Keep the content of the solution as-is and only insert [STEP] markers.

EXAMPLE:
{\rmfamily\itshape [Insert demonstration here]}

SOLUTION TO ANNOTATE:
<solution>
{\rmfamily\itshape [Insert reasoning trace here]}
</solution>

OUTPUT: Return the solution with all [STEP] markers inserted.
\end{prompttext}
\end{promptbox}

\newpage
We use the following demonstration throughout the experiments:

\begin{promptbox}
\begin{prompttext}
<solution>
First, let me understand the problem.
We need to find the value of x where 2x + 3 = 11.
Let's solve this by isolating x. Starting with 2x + 3 = 11, I subtract 3 from both sides to get 2x = 8. Then dividing by 2 gives x = 4.
```python
def f(x):
    return (11 - 3) / 2
```
Therefore, x is \textbackslash{}boxed\{\{4\}\}.
</solution>

<annotated_solution>
[STEP] First, let me understand the problem.
We need to find the value of x where 2x + 3 = 11.
[STEP] Let's solve this by isolating x. [STEP] Starting with 2x + 3 = 11, I subtract 3 from both sides to get 2x = 8. [STEP] Then dividing by 2 gives x = 4.
[STEP] ```python
def f(x):
    return (11 - 3) / 2
```
[STEP] Therefore, x is \textbackslash{}boxed\{\{4\}\}.
</annotated_solution>
\end{prompttext}
\end{promptbox}

\subsection{Example} \label{app:split-examples}

The following is a rejected greedy trace $\mathbf{t}_\texttt{AUG}$ generated and decomposed by Llama-3.1-8B-Instruct for a random problem from the OpenMathInstruct-2 dataset. Decomposing this trace based on simple rules such as line breaks would result in incomplete reasoning steps, whereas the model is capable of decomposing it into self-contained steps, using the decomposition prompt $\mathbf{p}_{\texttt{AUG}}$ from above:

\begin{reasoningbox}[title={\texttt{PROBLEM:} Let $a$ and $b$ be angles such that $\sin (a + b) = \frac{3}{5}$ and $\cos (a + b) = \frac{4}{5}$. Find
\[\tan \left( \frac{a + b}{2} \right).\] \\[-10pt]
\texttt{SOLUTION:} $\boxed{\frac{1}{3}}$}]
\small
\begin{reasonsteps}
\item To find $\tan \left( \frac{a + b}{2} \right)$, we can use the half-angle formula for tangent.

\item The half-angle formula for tangent is given by:

$\tan \left( \frac{\theta}{2} \right) = \frac{1 - \cos(\theta)}{\sin(\theta)}$

\item We are given that $\sin (a + b) = \frac{3}{5}$ and $\cos (a + b) = \frac{4}{5}$.

\item Using the angle sum identities for sine and cosine, we have:

$\sin(a+b) = \sin a \cos b + \cos a \sin b$

$\cos(a+b) = \cos a \cos b - \sin a \sin b$

\item We can rewrite these equations as:

$\sin a \cos b + \cos a \sin b = \frac{3}{5}$
$\cos a \cos b - \sin a \sin b = \frac{4}{5}$

\item Now, we can square both of these equations and add them together to get:

$(\sin a \cos b + \cos a \sin b)^2 + (\cos a \cos b - \sin a \sin b)^2 = \left(\frac{3}{5}\right)^2 + \left(\frac{4}{5}\right)^2$

\newpage
\item Expanding the left-hand side, we get:

$\sin^2 a \cos^2 b + 2 \sin a \cos a \sin b \cos b + \cos^2 a \sin^2 b + \cos^2 a \cos^2 b - 2 \sin a \cos a \sin b \cos b + \sin^2 a \sin^2 b = \frac{9}{25} + \frac{16}{25}$

\item Simplifying, we get:

$\sin^2 a \cos^2 b + \cos^2 a \sin^2 b + \cos^2 a \cos^2 b + \sin^2 a \sin^2 b = \frac{25}{25}$

$\sin^2 a (\cos^2 b + \sin^2 b) + \cos^2 a (\cos^2 b + \sin^2 b) = 1$

\item Since $\cos^2 b + \sin^2 b = 1$, we have:

$\sin^2 a + \cos^2 a = 1$

\item This is a Pythagorean identity, so it is true for all values of $a$.

\item Now, we can divide both sides of the equation by $\cos^2 a \cos^2 b$ to get:

$\frac{\sin^2 a}{\cos^2 a} \frac{1}{\cos^2 b} + \frac{\cos^2 a}{\cos^2 a} \frac{1}{\cos^2 b} = \frac{1}{\cos^2 a \cos^2 b}$

\item ...  %

\end{reasonsteps}
\end{reasoningbox}

\section{Reasoning step selection} \label{app:step-selection}

The analysis in \Cref{fig:uq} asks whether a ranking rule places useful alternative candidates early. For each rejected greedy trace from MATH-500, we generate one alternative candidate and its completed trace at every non-final reasoning step, without imposing a step budget. We then retrospectively rank the reasoning steps according to each selection score and measure the rank of the first accepted trace. This isolates selection from proposal generation.

The compared scores use greedy negative log-likelihood (G-NLL), inverse geometric mean token probability (PPL), the least probable token (MAX), reasoning step position alone (POS), or model self-reported confidence (LLM). Likelihood-based scores quantify local uncertainty \citep{aichberger2024rethi,fadeeva2023lmpolygraph}; position weighting favors earlier reasoning steps. The subscript POS denotes division by the one-indexed position $j$, with larger scores ranked first. The displayed LLM score uses confidence elicited with the reference answer.

The analysis is restricted to problems with both accepted and rejected completed traces (125 problems). If every completed trace of a problem is rejected, the rank of the first accepted trace is undefined; if every one is accepted, all scores yield rank one and cannot be distinguished. The mean rank is therefore conditional on recovery being possible in the recorded set, not an expected cost over all greedy-unsolved problems. It counts alternative candidates, not model calls, tokens, or wall-clock time.

\clearpage
\section{Reasoning step diversity sampling} \label{app:diversity-sampling}

\subsection{Prompt}
\vspace{-1pt}

\method{} and its three ablations in \Cref{sec:diversity-prompt} (\Cref{fig:ablation-math-500}) share the diversity prompt $\mathbf{p}_{\texttt{DIV}}$ shown below.
The prompt encourages the model to generate a self-contained alternative candidate $\mathbf{s}^{j,r}_{\texttt{ALT}}$ that follows the retained prefix $\mathbf{s}^{<j}$ and uses a different reasoning strategy from the selected reasoning step $\mathbf{s}^j$.
Blue text provides the reference answer and instructions for using it (\texttt{w/ sol}).
Green text requests a meta-thought before the alternative candidate (\texttt{w/ rsn}).
Removing either or both components gives the three ablations.
The \texttt{<thought>} and \texttt{<alternative\_step>} tags are used to extract the meta-thought and alternative candidate from the generated output.
The displayed prompt corresponds to the first proposal.
For subsequent proposals at the same position, the selected reasoning step and all previously generated alternative candidates are included as separate \texttt{<current\_step>} blocks, and the instruction asks for an alternative candidate that differs from all of them.
This changing context enables different proposals even at $T=0$.

\begin{promptbox}
\begin{prompttext}
You are given a problem that requires step-by-step reasoning.
You are also given previous steps that establish the reasoning so far.
To continue the reasoning, a potential current step has been proposed.
Your goal is to {\color{AppleGreen}first think of an *alternative* reasoning step that could best replace the current step and then} propose an *alternative* reasoning step that replaces the current step while remaining aligned with the previous steps.
{\color{AppleBlue}To achieve this goal more effectively, you are also given the solution to the problem. However, your proposed *alternative* reasoning step must not reveal or directly state this solution, but must advance the reasoning toward it.}

PROBLEM:
<problem>
{\color{AppleCoolGray}{\rmfamily\itshape [Problem statement]}}
</problem>

PREVIOUS STEPS:
<previous_steps>
{\color{AppleCoolGray}{\rmfamily\itshape [Previous reasoning steps]}}
</previous_steps>

CURRENT STEP:
<current_step>
{\color{AppleCoolGray}{\rmfamily\itshape [Selected reasoning step]}}
</current_step>

{\color{AppleBlue}SOLUTION:}
{\color{AppleBlue}<solution>}
{\color{AppleCoolGray}{\rmfamily\itshape [Reference answer]}}
{\color{AppleBlue}</solution>}

YOUR TASK:
Propose an *alternative* reasoning step that:
1. takes a fundamentally different conceptual approach compared to the current step above.
2. is a distinct and self-contained step, not a sequence or combination of steps.
3. serves as a direct continuation of the previous steps, phrased to naturally continue the reasoning in the same format and style.
4. replaces the current step, without repeating, rephrasing, or referring back to it.
5. is a smart but non-obvious approach, offering a fresh perspective that the current step does not anticipate.

{\color{AppleGreen}On a meta-level, first think strategically about the problem and which *alternative* reasoning step would best satisfy the requirements above. Provide these thoughts within <thought>...</thought> tags as a few concise sentences.}

Provide your *alternative* reasoning step within <alternative_step>...</alternative_step> tags as a single step.
\end{prompttext}
\end{promptbox}

\subsection{Example} \label{app:diversity-examples}

Consider the problem: given $\sin(a+b)=3/5$ and $\cos(a+b)=4/5$, find $\tan((a+b)/2)$. The reference answer is $1/3$. The example below shows Llama-3.1-8B-Instruct replacing the initial reasoning step $\mathbf{s}^1$ of the rejected trace in \Cref{app:split-examples} with the first alternative candidate $\mathbf{s}^{1,1}_{\texttt{ALT}}$, then generating the continuation $\mathbf{c}^{1,1}$. The meta-thought guides the proposal but is excluded from the completed trace.

\begin{reasoningbox}[title={\texttt{CURRENT STEP:} To find $\tan \left( \frac{a + b}{2} \right)$, we can use the half-angle formula for tangent.}]

\texttt{META-THOUGHT AND ALTERNATIVE CANDIDATE:}
\vspace{4pt}
\begin{spacing}{1.2}
\small

<thought>

Given the information provided, an alternative reasoning step should focus on leveraging the given trigonometric values to directly compute the tangent of the half-angle, rather than relying on a formula. This could involve using trigonometric identities or properties that relate the tangent of a sum of angles to the tangent of its half-angle. By taking a different approach, we may uncover a more elegant or insightful solution.

</thought>

<alternative\_step>

Let's reconsider the relationship between tangent and the sine and cosine functions. We know that $\tan \theta = \frac{\sin \theta}{\cos \theta}$. Given that $\sin (a + b) = \frac{3}{5}$ and $\cos (a + b) = \frac{4}{5}$, we can express the tangent of the sum of angles $a$ and $b$ as $\tan (a + b) = \frac{\sin (a + b)}{\cos (a + b)} = \frac{\frac{3}{5}}{\frac{4}{5}} = \frac{3}{4}$.

</alternative\_step>
\end{spacing}

\vspace{4pt}
\noindent\hrulefill
\vspace{4pt}

\vspace{4pt}
\texttt{CONTINUATION:} 
\vspace{4pt}
\begin{spacing}{1.2}
\small
Now, let's explore how this relates to the tangent of the half-angle $\frac{a + b}{2}$.

\vspace{4pt}
We can use the half-angle formula for tangent to find $\tan \left( \frac{a + b}{2} \right)$:

$\tan \left( \frac{a + b}{2} \right) = \frac{1 - \cos (a + b)}{\sin (a + b)}$

\vspace{4pt}
Substituting the given values, we get:

$\tan \left( \frac{a + b}{2} \right) = \frac{1 - \frac{4}{5}}{\frac{3}{5}}$

\vspace{4pt}
Simplifying the expression, we have:

$\tan \left( \frac{a + b}{2} \right) = \frac{\frac{1}{5}}{\frac{3}{5}}$

\vspace{4pt}
Now, let's simplify the fraction:

$\tan \left( \frac{a + b}{2} \right) = \frac{1}{3}$

\vspace{4pt}
Therefore, the value of $\tan \left( \frac{a + b}{2} \right)$ is $\boxed{\frac{1}{3}}$.

\vspace{-10pt}
\end{spacing}
\end{reasoningbox}

\newpage
\subsection{Ablation} \label{app:diversity-ablation}

\begin{figure*}[htbp]
    \centering
    \includegraphics[width=\linewidth,trim=0cm 0.45cm 0.35cm 0.39cm,clip]{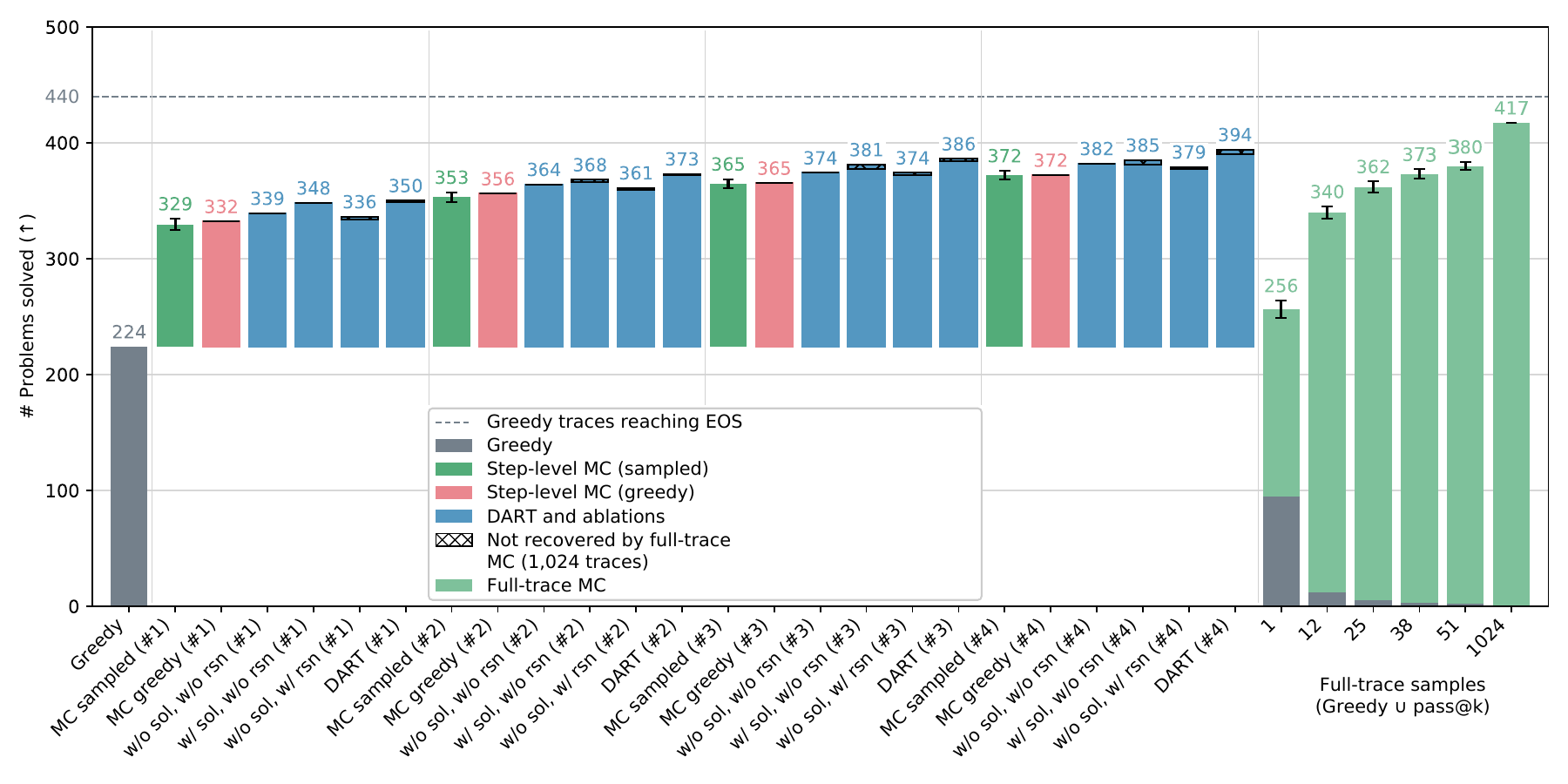}
    \caption{MATH-500 coverage across sampling strategies. All bars include the 224 greedy-solved problems (gray); completed traces are generated from rejected greedy traces that terminate normally. Labels \#1--4 denote alternative candidates per selected reasoning step. Dark green denotes step-level MC sampling with sampled continuation, red denotes step-level MC sampling with greedy continuation, and blue denotes \method{} and its three ablations with greedy continuation. Light green denotes full-trace MC sampling from scratch; MC sampling uses $T=1$. The full-trace axis reports Greedy $\cup$ pass@$k$. Error bars show one standard deviation of stochastic-only coverage under resampling, before including separate greedy-solved problems. Hatched portions mark problems not recovered by full-trace MC sampling with 1,024 traces. The dashed line at 440 marks the number of normally terminated greedy traces.}
    \label{fig:app_ablation-math-500}
\end{figure*}

\Cref{fig:app_ablation-math-500} extends \Cref{fig:ablation-math-500} from one to four alternative candidates per selected reasoning step, with one completed trace per alternative candidate. At four alternative candidates, \method{} covers 394 problems, including 224 greedy-solved problems, compared with 372 for step-level MC sampling with greedy continuation. The difference is 22 recovered problems, or 14.9\% more recoveries among greedy-unsolved problems. Full-trace MC sampling with the largest displayed number of traces covers 417 problems but still misses some recovered by \method{}. The ablation therefore supports complementary coverage, not dominance at arbitrary budgets.

\clearpage
\section{Human and model-adjusted difficulty}
\label{app:inference}

\paragraph{Model-adjusted strata.}
Difficulty levels 1 through 8 use the inclusive ranges $[135,512]$, $[85,133]$, $[52,83]$, $[32,50]$, $[17,30]$, $[7,15]$, $[1,4]$, and $\{0\}$ for the number of accepted traces among 512 independently sampled traces, given by $512 \widehat  q(\mathbf{x})$ (\Cref{eq:model-adjusted-difficulty}). 
These ranges were selected to obtain approximately equally sized difficulty strata.
The implementation traverses aligned records of \method{} and MC sampling in stored order and retains up to the first 255 greedy-unsolved problems in each range. The methods are compared on identical strata, not on every greedy-unsolved problem. This subsampling affects neither trace generation nor post-training.

For a budget of $k$ completed traces, MC sampling uses entries 401 through $400+k$ of the same ordered pool used to estimate $\widehat q(\mathbf{x})$. It is therefore conditional on the pool defining each stratum. In particular, by construction, MC sampling recovers no problem in the stratum without accepted traces. An independent difficulty-estimation pool would be needed to evaluate recovery without this dependence.

\begin{figure*}[p]
    \centering
    {\small
    \begin{tabular*}{\textwidth}{@{\extracolsep{\fill}}lll@{}}
        \colorswatch{AppleBlue5}~~\method{}, first selected reasoning step &
        \colorswatch{AppleBlue3}~~\method{}, additional selected reasoning steps \\
        \colorswatch{AppleGreen4}~~both methods &
        \crossedswatch{AppleRed4}~~\method{} only &
        \colorswatch{AppleRed4}~~MC sampling only
        \vspace{8pt}
    \end{tabular*}}
    \includegraphics[trim={0.5cm 0 0.5cm 2.1cm},clip,width=\linewidth]{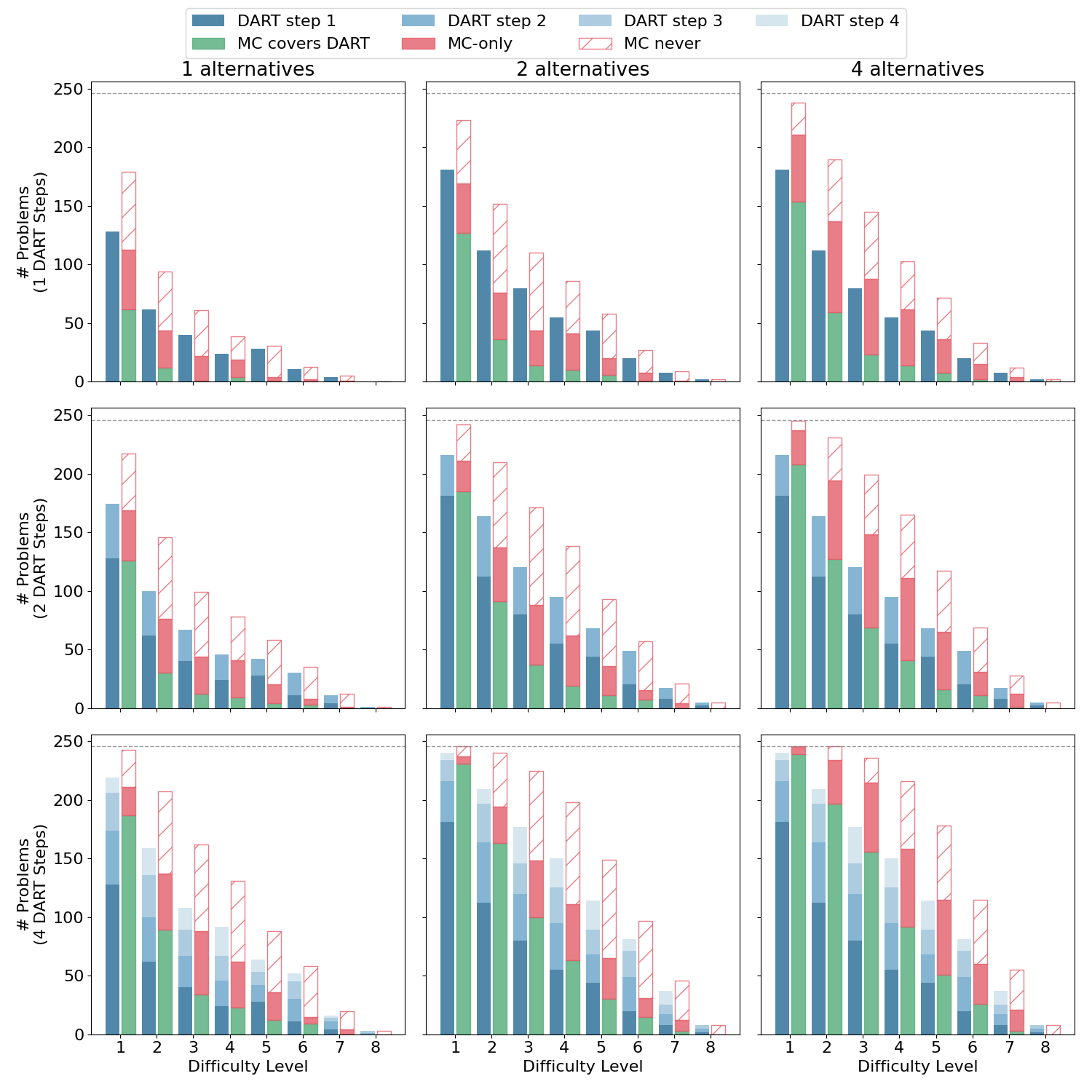}
    \caption{OpenMathInstruct-2 recovery by model-adjusted difficulty. All problems are greedy-unsolved. Rows use one, two, or four selected reasoning steps. Columns use one, two, or four alternative candidates per reasoning step. Blue shades denote cumulative \method{} recoveries, green denotes recovery by both methods, solid red denotes recovery by MC sampling only, and red hatching denotes \method{}-only recovery at the matched completed-trace budget. The traces from MC sampling also contribute to the difficulty estimate $\widehat q(\mathbf{x})$.}
    \label{fig:open_math_difficulty}
\end{figure*}

\begin{figure*}[p]
    \centering
    \centering
    {\small
    \begin{tabular*}{\textwidth}{@{\extracolsep{\fill}}lll@{}}
        \colorswatch{AppleBlue5}~~\method{}, first selected reasoning step &
        \colorswatch{AppleBlue3}~~\method{}, additional selected reasoning steps \\
        \colorswatch{AppleGreen4}~~both methods &
        \crossedswatch{AppleRed4}~~\method{} only &
        \colorswatch{AppleRed4}~~MC sampling only
        \vspace{8pt}
    \end{tabular*}}
    \includegraphics[trim={0.5cm 0 0.5cm 2.1cm},clip,width=\linewidth]{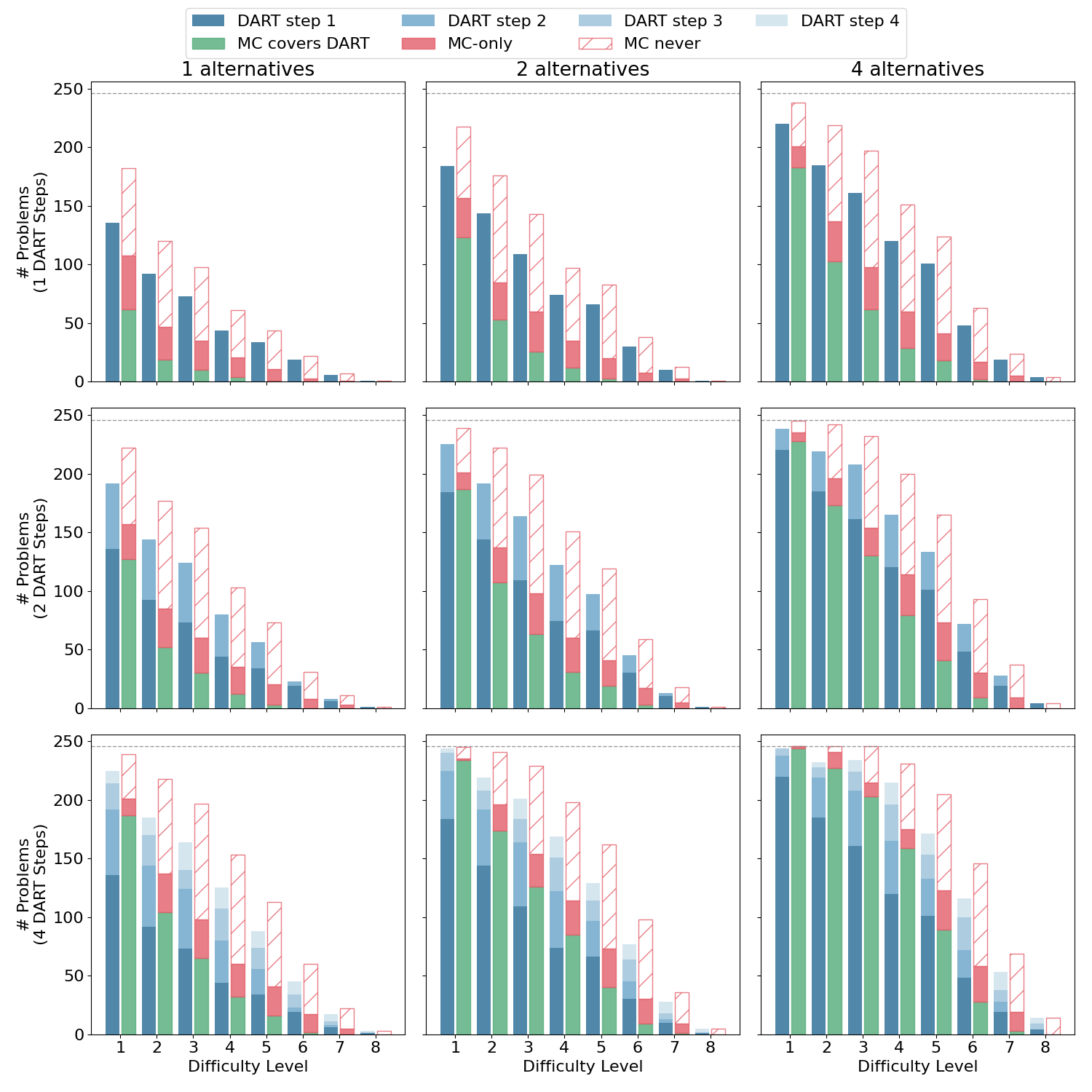}
    \caption{DeepMath-103K recovery by model-adjusted difficulty, using the same layout, completed-trace budgets, and colors as \Cref{fig:open_math_difficulty}. All problems are greedy-unsolved. The traces from MC sampling also contribute to $\widehat q(\mathbf{x})$, so zero recovery by MC sampling in the hardest stratum follows from its definition.}
    \label{fig:deep_math_difficulty}
\end{figure*}

\Cref{fig:open_math_difficulty,fig:deep_math_difficulty} extend \Cref{fig:difficulty_small} to one, two, or four selected reasoning steps and one, two, or four alternative candidates per reasoning step. Their combinations yield budgets up to $k=16$, with one completed trace per alternative candidate. Blue shades separate recoveries by the first selected reasoning step that yields an accepted trace. The adjacent bar separates overlap with MC sampling from method-specific recoveries.

\begin{figure*}[p]
    \centering
    \centering
    {\small
    \begin{tabular*}{\textwidth}{@{\extracolsep{\fill}}lll@{}}
        \colorswatch{AppleBlue5}~~\method{}, first selected reasoning step &
        \colorswatch{AppleBlue3}~~\method{}, additional selected reasoning steps &
        \colorswatch{AppleCoolGray}~~Greedy \\
        \colorswatch{AppleGreen4}~~both methods &
        \crossedswatch{AppleRed4}~~\method{} only &
        \colorswatch{AppleRed4}~~MC sampling only
        \vspace{8pt}
    \end{tabular*}}
    \includegraphics[trim={0.5cm 0 0.5cm 2.1cm},clip,width=\linewidth] {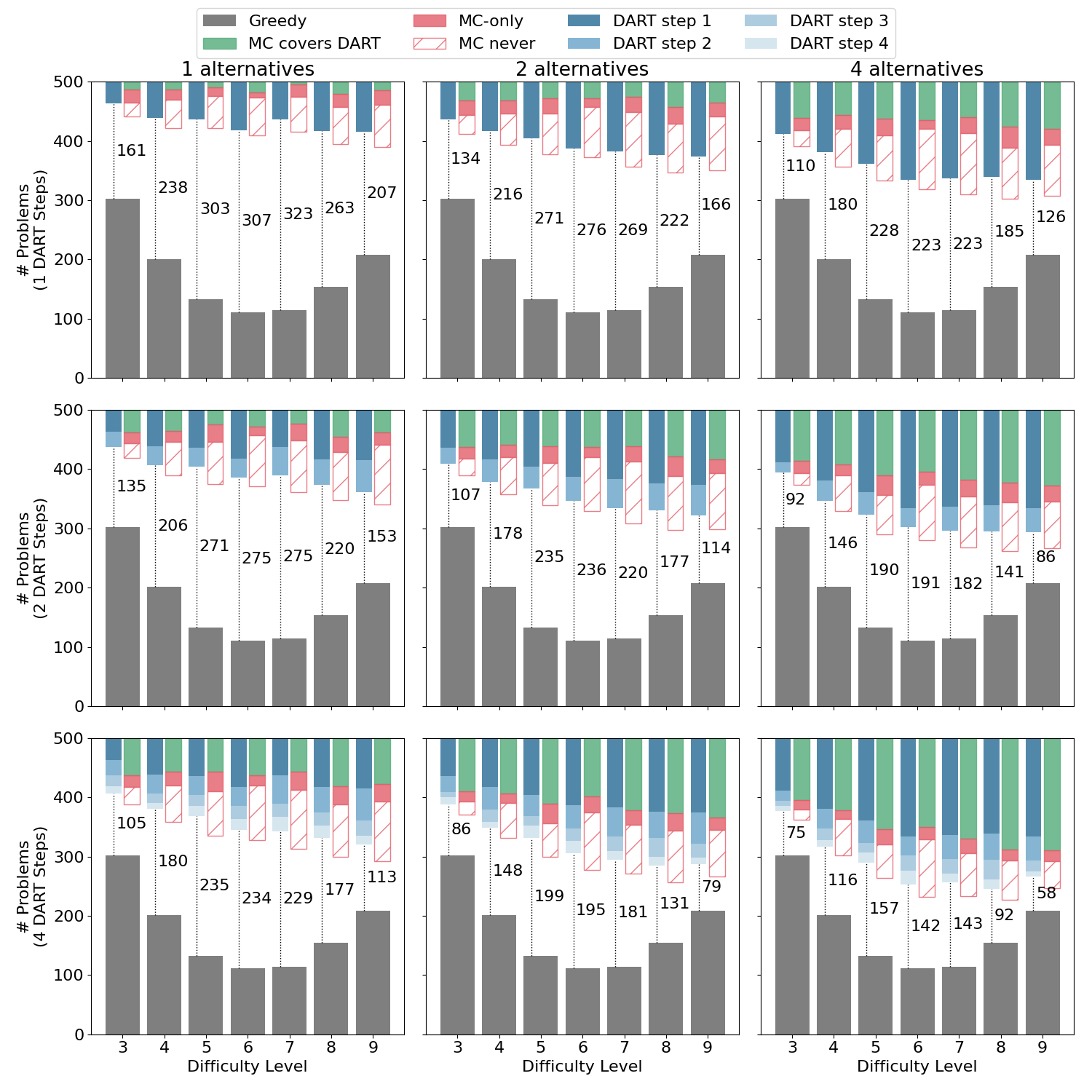}
    \caption{DeepMath-103K problem counts grouped by assigned AoPS difficulty. Gray denotes greedy-solved problems. The remaining bars compare recovery among greedy-unsolved problems: blue shades denote cumulative \method{} recoveries across selected reasoning steps, green denotes recovery by both methods, solid red denotes recovery by MC sampling only, and red hatching denotes \method{}-only recovery at the displayed budget. Rows use one, two, or four selected reasoning steps. Columns use one, two, or four alternative candidates per reasoning step. Annotations next to the dotted line count problems not covered by greedy decoding or \method{}.}
    \label{fig:deepmath_original_difficulty}
\end{figure*}

\paragraph{Human-oriented labels.}
\Cref{fig:deepmath_original_difficulty} groups the DeepMath-103K subset by its GPT-4o annotations based on Art of Problem Solving (AoPS) guidelines. The number of greedy-solved problems is not monotone in the assigned level: levels 8--9 contain more than levels 5--7 in this subset. These counts depend on the number of problems at each level and do not establish a reversal in per-problem success probability. Human-oriented curricular labels and the model-dependent quantity $q_{\boldsymbol{\theta}}(\mathbf{x})$ serve different purposes. We use $\widehat q(\mathbf{x})$ to describe recovery within the recorded pools of MC sampling.

\clearpage
\section{Generation-cost accounting}
\label{app:compute-accounting}

The analysis in \Cref{fig:compute-ablation} estimates algorithmic matrix-multiplication FLOPs rather than accelerator utilization or elapsed time. For method $m$, let $T_m$ denote the sum of recorded uncached prompt tokens and generated tokens, $P_m$ the number of cache-aware causal token pairs, and $Z_m$ the number of vocabulary-head evaluations. For a cached prefix of length $c$ followed by an uncached prefill suffix of length $u$, the prefill contributes $cu+u(u+1)/2$ pairs. Generating $d$ reported tokens after a prompt of length $p$ contributes $pd+d(d+1)/2$ pairs. Thus, the telemetry convention treats every reported generated token as a decode-query position and can overcount transformer evaluations by one position per returned sequence. If one request returns multiple sequences, its first-token logits are shared; hence $Z_m=G_m-S_m+R_m$, where $G_m$, $S_m$, and $R_m$ are the numbers of generated tokens, returned sequences, and requests, respectively.

For a dense gated decoder with $L$ layers, hidden width $h$, intermediate width $f$, query width $d_Q$, key/value width $d_{KV}$, and vocabulary size $V$, we compute
\begin{equation}
\begin{split}
F_m={}&2LT_m\bigl(hd_Q+2hd_{KV}+d_Qh+3hf\bigr)\\
&+4Ld_QP_m+2hVZ_m.
\end{split}
\label{eq:compute-flops}
\end{equation}
The first term counts Q/K/V/O and gated-MLP projections, the second counts the query-key and attention-value products, and the third counts the vocabulary projection. We count each multiply-add as two FLOPs. For Llama-3.1-8B-Instruct, $L=32$, $h=d_Q=4{,}096$, $d_{KV}=1{,}024$, $f=14{,}336$, and $V=128{,}256$. These dimensions give 13,958,643,712 projection and MLP FLOPs per model token, 524,288 attention FLOPs per causal token pair, and 1,050,673,152 FLOPs per vocabulary-head evaluation. \Cref{tab:compute-counters} reports the resulting counters and estimates.

The estimate charges only uncached prompt queries while retaining the attention work of decoded queries over cached keys. It excludes embedding lookup, normalization, RoPE, nonlinear activations, softmax, sampling, padding, communication, verifier and CPU work, and any extra operations executed by a particular kernel implementation. The shared greedy generation, reasoning step decomposition, training, and evaluation are also outside the comparison. Consequently, \Cref{fig:compute-ablation} estimates the dominant generation operations after reasoning step selection, not end-to-end runtime, energy, or hardware cost.

\begin{table}[h]
    \centering
    \small
    \caption{Recorded counters and FLOP estimates for the 6,200 shared greedy-unsolved problems. \emph{Recovered} denotes the number of problems with at least one accepted trace and \emph{Accepted} denotes the total number of accepted traces. $T_m$ and $Z_m$ are in millions, $P_m$ in billions, and $F_m$ (\Cref{eq:compute-flops}) in units of $10^{18}$ FLOPs.}
    \label{tab:compute-counters}
    \vspace{4pt}
    \renewcommand{\arraystretch}{1.3}
    \setlength{\tabcolsep}{13.2pt}
    \begin{tabular}{lrrrrrr}
        \toprule
        Method & $T_m$ & $P_m$ & $Z_m$ & Recovered & Accepted & $F_m$ \\
        \midrule
        MC sampling & 86.85 & 145.26 & 86.73 & 2,168 & 3,987 & 1.380 \\
        \method{} & 124.52 & 158.36 & 90.81 & 2,971 & 7,921 & 1.917 \\
        \bottomrule
    \end{tabular}
\end{table}

\clearpage
\section{Post-training protocol}
\label{app:posttraining-protocol}

\paragraph{Data.}
We mine traces from 16,384 OpenMathInstruct-2 training problems or 8,192 DeepMath-103K training problems, filtering empty answers and exact duplicate problem strings. DeepMath-103K problems have annotated difficulty at least five. We evaluate on MATH-500 and the 2022--2025 American Invitational Mathematics Examinations (AIME) \citep{maaAIME}. No dedicated near-duplicate decontamination procedure was run beyond exact problem-string deduplication, so residual overlap or benchmark contamination cannot be excluded.

\paragraph{Trace mining.}
All mining methods use Llama-3.1-8B-Instruct. Greedy traces use $T=0$; MC sampling and RFT use $T=1$. Reasoning traces are limited to 4,096 tokens, and generation uses a repetition penalty of 1.05. For \method{}, the procedure in \Cref{app:split-prompt} decomposes each rejected greedy trace, allowing ten closed reasoning steps plus any remaining suffix. We select up to four non-final reasoning steps and generate two alternative candidates per selected reasoning step. The meta-thought, alternative candidate, and continuation each use $T=0$. Alternative candidates are generated sequentially, with earlier alternative candidates included in the next diversity prompt as described in \Cref{app:diversity-sampling}. The reference answer is provided to \method{}, STaR, and RFT prompts. Only accepted traces are retained, and accepted greedy traces are included in every training corpus. Meta-thoughts guide the proposals but are excluded from the training corpora.

\paragraph{Generation stages.}
For each alternative candidate, \method{} generates a meta-thought, the alternative candidate itself, and a continuation.
The meta-thought and alternative candidate can be generated jointly, and generation can be batched across reasoning steps and problems.
Retained prefixes are not decoded again, but their context can incur prefill work unless KV caches are reused.
We conduct a dedicated generation-cost analysis using a separately instrumented configuration (\Cref{app:compute-accounting}).

\paragraph{Supervised fine-tuning.}
Within each mining dataset, all runs start from the base checkpoint and use AdamW for two epochs, learning rate $10^{-5}$, weight decay 0, 10\% linear warm-up, effective batch size 128, maximum sequence length 4,096, and seed 42. The mining method determines the number and composition of retained traces; training corpora are not matched in examples or tokens.

\paragraph{Evaluation and intervals.}
Greedy $\cup$ pass@$k$ counts an evaluation problem as solved if a separate greedy trace or at least one of $k$ stochastic traces is accepted. Stochastic traces are generated at $T=1$. For each resampling draw, we select $k$ traces without replacement from each problem's recorded pool and compute accuracy across the fixed evaluation set. We report the mean across draws and the central 95\% empirical range. These intervals are conditional on the evaluation problems, recorded traces, and selected checkpoint; they do not resample problems or rerun generation or training. \Cref{tab:finetuning-main} uses $k=32$ and reports the highest observed point estimate across evaluated checkpoints on each displayed evaluation set. These are exploratory, evaluation-selected results.

\begin{figure*}[p]
    \centering
    \begin{subfigure}[t]{\linewidth}
        \centering
        \includegraphics[width=\linewidth, trim=0cm 0.5cm 0cm 2cm, clip]{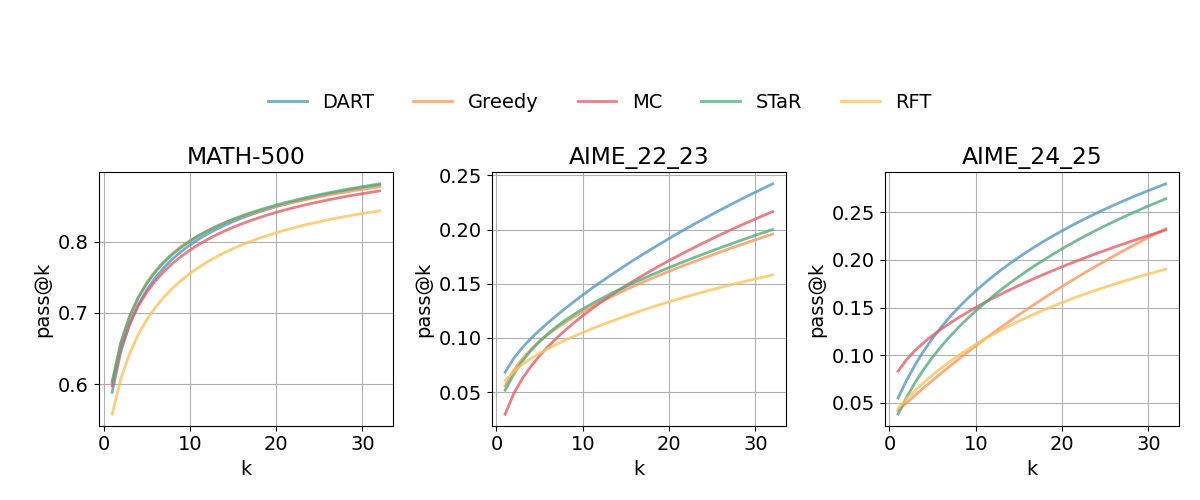}
        \caption{Checkpoint selected separately on each evaluation set.}
        \label{fig:best_pass_at_k_overall}
    \end{subfigure}

    \vspace{0.1cm}
    \begin{subfigure}[t]{\linewidth}
        \centering
        \includegraphics[width=\linewidth, trim=0cm 0.5cm 0cm 3cm, clip]{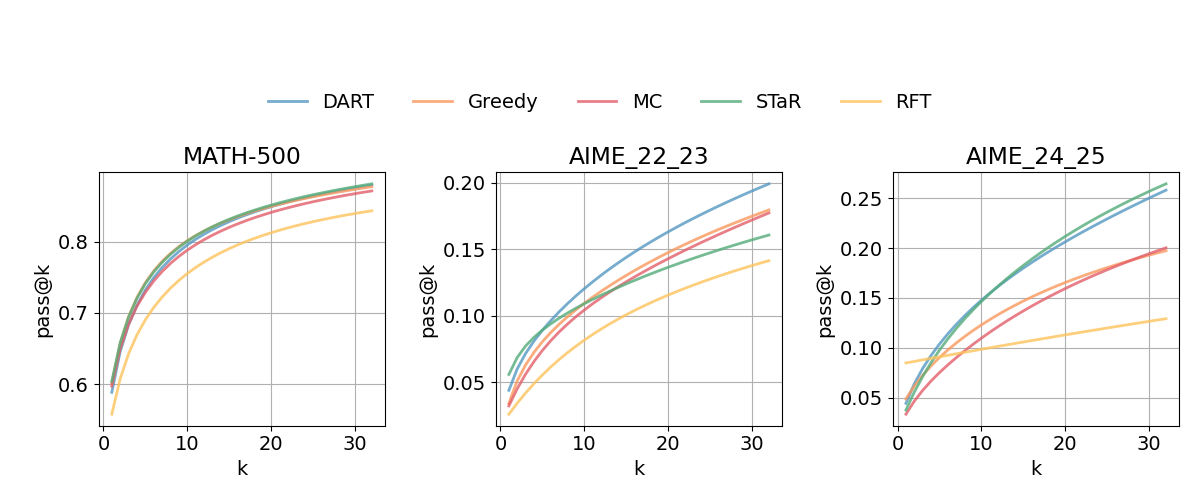}
        \caption{Checkpoint selected on MATH-500.}
        \label{fig:best_pass_at_k_math500}
    \end{subfigure}

    \vspace{0.1cm}
    \begin{subfigure}[t]{\linewidth}
        \centering
        \includegraphics[width=\linewidth, trim=0cm 0.5cm 0cm 3cm, clip]{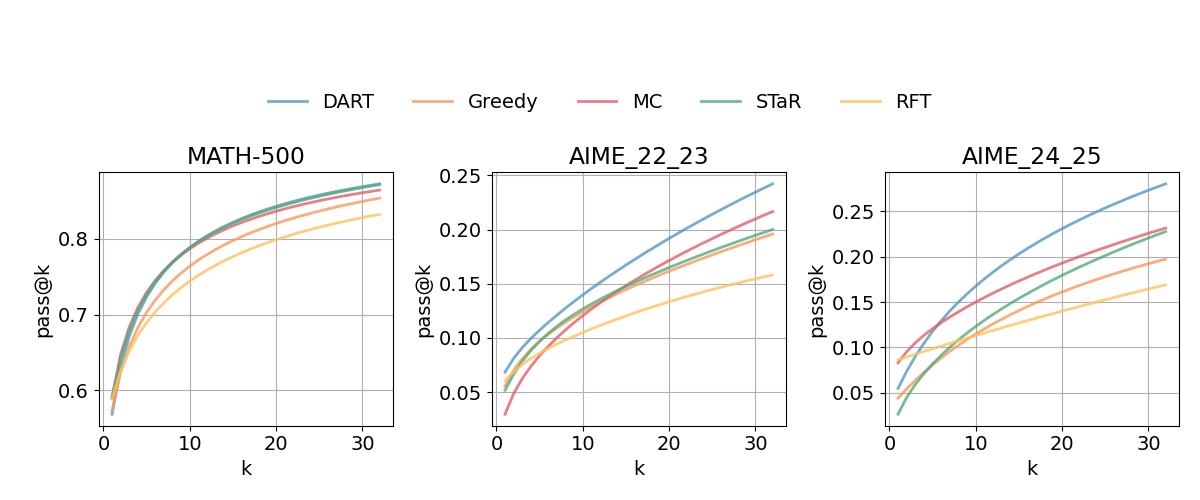}
        \caption{Checkpoint selected on AIME 22+23.}
        \label{fig:best_pass_at_k_aime2223}
    \end{subfigure}

    \vspace{0.1cm}
    \begin{subfigure}[t]{\linewidth}
        \centering
        \includegraphics[width=\linewidth, trim=0cm 0.5cm 0cm 3cm, clip]{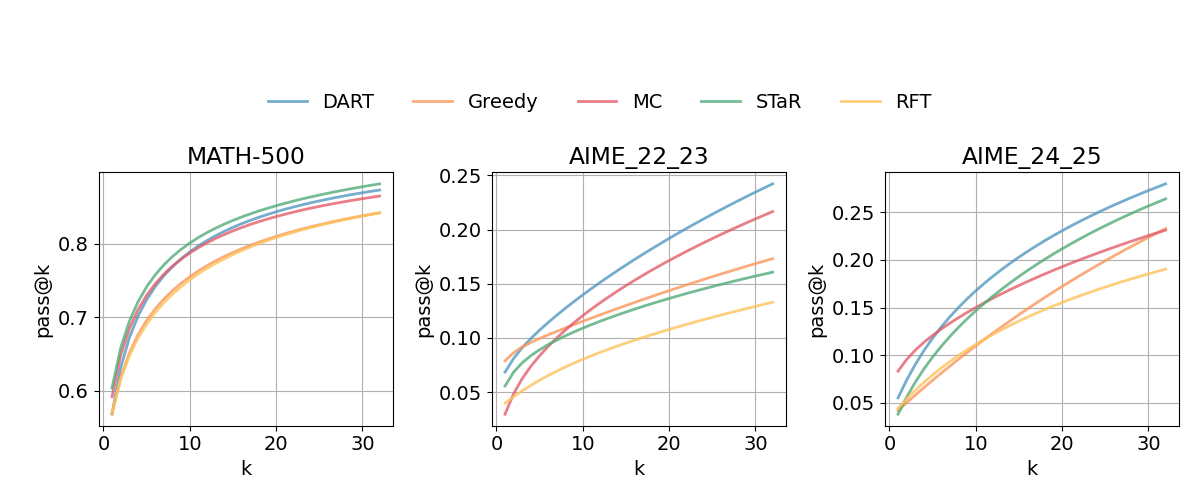}
        \caption{Checkpoint selected on AIME 24+25.}
        \label{fig:best_pass_at_k_aime2425}
    \end{subfigure}

    \caption{Greedy $\cup$ pass@$k$ accuracy of Llama-3.1-8B-Instruct after SFT on traces mined from 16k OpenMathInstruct-2 problems. Plot labels pass@$k$ denote this union metric, and STaR and RFT denote the adapted mining methods in \Cref{sec:post-training}. Checkpoints are selected by Greedy $\cup$ pass@32 on the set named in each panel.}
    \label{fig:best_pass_at_k}
\end{figure*}

\paragraph{Sampling-budget curves.}
\Cref{fig:best_pass_at_k} reports Greedy $\cup$ pass@$k$ for post-training on OpenMathInstruct-2. Panel (\subref{fig:best_pass_at_k_overall}) selects a checkpoint separately on each evaluation set. Panels (\subref{fig:best_pass_at_k_math500}--\subref{fig:best_pass_at_k_aime2425}) select one checkpoint per mining method using the named evaluation set, then evaluate it on all three sets. This distinguishes per-set checkpoint selection from cross-set evaluation of a fixed checkpoint; it does not remove the selection bias on the set used to choose that checkpoint.

\clearpage
\section{Broader impacts and existing assets}
\label{app:disclosures}

\subsection{Broader impacts}
\label{app:broader-impacts}

\method{} improves the efficiency and coverage of reasoning-trace generation, supporting the development of more capable language models for mathematical reasoning and related tasks. However, final-answer verification does not guarantee the correctness of intermediate reasoning steps, and training on accepted traces may propagate such errors. More broadly, advances in reasoning capabilities may have implications beyond the mathematical domains studied here. Our experiments focus on mathematical reasoning and do not evaluate these broader effects.

\subsection{Existing assets and licenses}
\label{app:asset-licenses}

\Cref{tab:asset-licenses} identifies the principal model and data assets, their roles, and the licenses stated by their maintainers. We cite both the MATH benchmark \citep{hendrycksMeasuringMathematicalProblem2021} and the work introducing its 500-problem evaluation split \citep{lightmanLetsVerifyStep2023}. Upstream repository licenses do not establish separate rights in all underlying competition content. In particular, the public \href{https://huggingface.co/datasets/allenai/aime-2022-2025}{Ai2 AIME dataset card} does not state a license; we do not infer one from its public availability. AIME is used for evaluation, and no copy of this dataset is redistributed with the submission.

The implementation uses \href{https://github.com/pytorch/pytorch/blob/main/LICENSE}{PyTorch (BSD-style)}, \href{https://github.com/vllm-project/vllm/blob/main/LICENSE}{vLLM (Apache 2.0)}, \href{https://github.com/ray-project/ray/blob/master/LICENSE}{Ray (Apache 2.0)}, and \href{https://github.com/deepspeedai/DeepSpeed/blob/master/LICENSE}{DeepSpeed (Apache 2.0)}. Hugging Face's \href{https://github.com/huggingface/transformers/blob/main/LICENSE}{Transformers}, \href{https://github.com/huggingface/datasets/blob/main/LICENSE}{Datasets}, \href{https://github.com/huggingface/trl/blob/main/LICENSE}{TRL}, and \href{https://github.com/huggingface/Math-Verify/blob/main/LICENCE}{Math-Verify} each use Apache 2.0. The links credit the software projects and provide their license texts. We do not redistribute these packages, source datasets, generated training corpora, or model checkpoints with the submission.

\begin{table}[h]
    \centering
    \small
    \caption{Principal existing model and data assets. Links identify their upstream repositories and license information; no new license is assigned to these assets.}
    \vspace{4pt}
    \renewcommand{\arraystretch}{1.3}
    \label{tab:asset-licenses}
    \begin{tabular}{@{}p{0.40\linewidth}p{0.57\linewidth}@{}}
        \toprule
        Asset and source & Role and stated license \\
        \midrule
        \href{https://huggingface.co/meta-llama/Llama-3.1-8B-Instruct}{Llama-3.1-8B-Instruct} \citep{grattafiori2024}
        & Trace generation and SFT initialization. \href{https://github.com/meta-llama/llama-models/blob/main/models/llama3_1/LICENSE}{Llama 3.1 Community License}, including its Acceptable Use Policy. \\[4pt]
        \href{https://huggingface.co/datasets/zwhe99/DeepMath-103K}{DeepMath-103K} \citep{he2025deepmath}
        & Mining problems and reference answers from the training split. Dataset card: MIT. \\[4pt]
        \href{https://huggingface.co/datasets/nvidia/OpenMathInstruct-2}{OpenMathInstruct-2} \citep{toshniwal2024openmathinstruct}
        & Mining problems and reference answers from \texttt{train\_1M}. Dataset card: CC BY 4.0. \\[4pt]
        \href{https://github.com/openai/prm800k}{MATH-500} \citep{lightmanLetsVerifyStep2023}
        & Method development and evaluation. The \href{https://github.com/hendrycks/math/blob/main/LICENSE}{MATH} and \href{https://github.com/openai/prm800k/blob/main/LICENSE}{PRM800K} repositories each state MIT. \\[4pt]
        \href{https://huggingface.co/datasets/allenai/aime-2022-2025}{AIME 2022--2025, Ai2 compilation} \citep{maaAIME}
        & Evaluation on MAA competition problems. No license stated on the dataset card. \\
        \bottomrule
    \end{tabular}
\end{table}

\clearpage

\end{document}